\documentclass[10pt,twocolumn,letterpaper]{article}

\usepackage{cvpr}
\usepackage{times}
\usepackage{epsfig}
\usepackage{graphicx}
\usepackage{amsmath}
\usepackage{amssymb}

\cvprfinalcopy

\def\cvprPaperID{3440}

\ifcvprfinal\fi
\usepackage{amsmath}
\usepackage{nicefrac}
\usepackage{amsthm}
\usepackage{xcolor}
\usepackage{subcaption}
\usepackage[inline]{enumitem}
\usepackage{xspace}
\usepackage{tabularx}
\newcolumntype{R}{>{\raggedleft\arraybackslash}X}

\definecolor{colorbrewer0}{RGB}{45,45,45}
\definecolor{colorbrewer1}{RGB}{228,26,28}
\definecolor{colorbrewer2}{RGB}{55,126,184}
\definecolor{colorbrewer3}{RGB}{77,175,74}
\definecolor{colorbrewer4}{RGB}{152,78,163}
\definecolor{colorbrewer5}{RGB}{255,127,0}
\definecolor{colorbrewer6}{RGB}{255,255,51}
\definecolor{colorbrewer7}{RGB}{166,86,40}
\definecolor{colorbrewer8}{RGB}{247,129,191}
\definecolor{colorbrewer9}{RGB}{153,153,153}

\newcommand{\mR}{\mathbb{R}}

\newcommand{\mE}{\mathbb{E}}

\newcommand{\cL}{\mathcal{L}}

\newcommand{\figref}[1]{\Fig~\ref{#1}}
\newcommand{\secref}[1]{Section~\ref{#1}}
\newcommand{\defref}[1]{Def.~\ref{#1}}
\newcommand{\eqnref}[1]{Eq.~\eqref{#1}}

\DeclareMathOperator*{\argmin}{argmin~}

\DeclareMathOperator{\enc}{enc}
\DeclareMathOperator{\dec}{dec}
\DeclareMathOperator{\dis}{dis}
\newcommand{\inner}[1]{\left\langle#1\right\rangle}

\makeatletter
\DeclareRobustCommand\onedot{\futurelet\@let@token\@onedot}
\def\@onedot{\ifx\@let@token.\else.\null\fi\xspace}
\def\eg{e.g\onedot} 
\def\ie{i.e\onedot} 
\def\cf{cf\onedot} 
\def\etc{etc\onedot} \def\vs{vs\onedot}
\def\wrt{wrt\onedot}

\def\etal{et~al\onedot} 
\def\Fig{Fig\onedot}   
\makeatother

\newtheoremstyle{small}
  {2px} 
  {2px} 
  {} 
  {} 
  {\bfseries} 
  {.} 
  {.5em} 
  {} 
\theoremstyle{small}\newtheorem{definition}{Definition}
\theoremstyle{small}
\theoremstyle{small}\newtheorem*{hypothesis*}{Hypothesis}

\newcommand{\Fonts}{FONTS\xspace}
\newcommand{\MNIST}{EMNIST\xspace}
\newcommand{\Fashion}{F-MNIST\xspace}
\newcommand{\FashionMNIST}{Fashion-MNIST\xspace}
\newcommand{\Celeb}{CelebA\xspace}

\newcommand{\VAEGAN}{VAE-GAN\xspace}
\newcommand{\VAEGANs}{VAE-GANs\xspace}

\newcommand{\matthias}[1]{#1}
\newcommand{\myparagraph}[1]{\vspace{3pt}\noindent{\bf #1}}
\newcommand{\red}[1]{#1}
\usepackage[pagebackref=true,breaklinks=true,letterpaper=true,colorlinks,bookmarks=false]{hyperref}

\begin{document}

\title{Disentangling Adversarial Robustness and Generalization}
\author{David Stutz$^{1}$ \qquad Matthias Hein$^{2}$ \qquad Bernt Schiele$^{1}$\\
$^1$Max Planck Institute for Informatics, Saarland Informatics Campus, Saarbr\"{u}cken\\
$^3$University of T\"{u}bingen, T\"{u}bingen\\
{\tt\small \{david.stutz,schiele\}@mpi-inf.mpg.de, matthias.hein@uni-tuebingen.de}	
}

\maketitle
\thispagestyle{empty}

\begin{abstract}
    Obtaining deep networks that are robust against adversarial examples \emph{and} generalize well is an open problem. A recent hypothesis \cite{TsiprasARXIV2018,SuARXV2018} even states that \emph{both} robust \emph{and} accurate models are impossible, \ie, adversarial robustness and generalization are conflicting goals. In an effort to clarify the relationship between robustness and generalization, we assume an underlying, low-dimensional data manifold and show that:
	\begin{enumerate*}
    	\item regular adversarial examples leave the manifold;
    	\item adversarial examples constrained to the manifold, \ie, on-manifold adversarial examples, exist;
    	\item on-manifold adversarial examples are generalization errors, and on-manifold adversarial training boosts generalization;
    	\item \red{regular robustness and generalization are not necessarily contradicting goals.}
	\end{enumerate*}
    These assumptions imply that \emph{both} robust \emph{and} accurate models are possible. However, different models (architectures, training strategies \etc) can exhibit different robustness and generalization characteristics. To confirm our claims, we present extensive experiments on synthetic data (with known manifold) as well as on \MNIST \cite{CohenARXIV2017}, \FashionMNIST \cite{XiaoARXIV2017} and \Celeb \cite{LiuICCV2015}.
\end{abstract}
\vspace*{-0.4cm}
\section{Introduction}
\label{sec:introduction}

\begin{figure}
    \centering
    \vskip -0.25cm
    \begin{subfigure}[t]{0.5\textwidth}
        \centering
        \hspace*{-0.5cm}
        \includegraphics[width=0.9\textwidth]{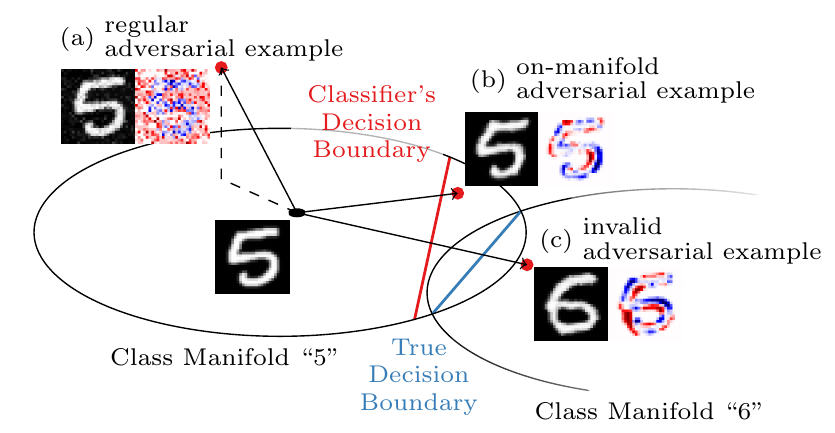}
    \end{subfigure}
    \vskip -8px
    {\color{black!75}\noindent\rule{0.45\textwidth}{0.4pt}}
    \vskip 4px
    \begin{subfigure}[t]{0.5\textwidth}
        \centering
        \hskip -0.5cm
        \includegraphics[width=0.8\textwidth]{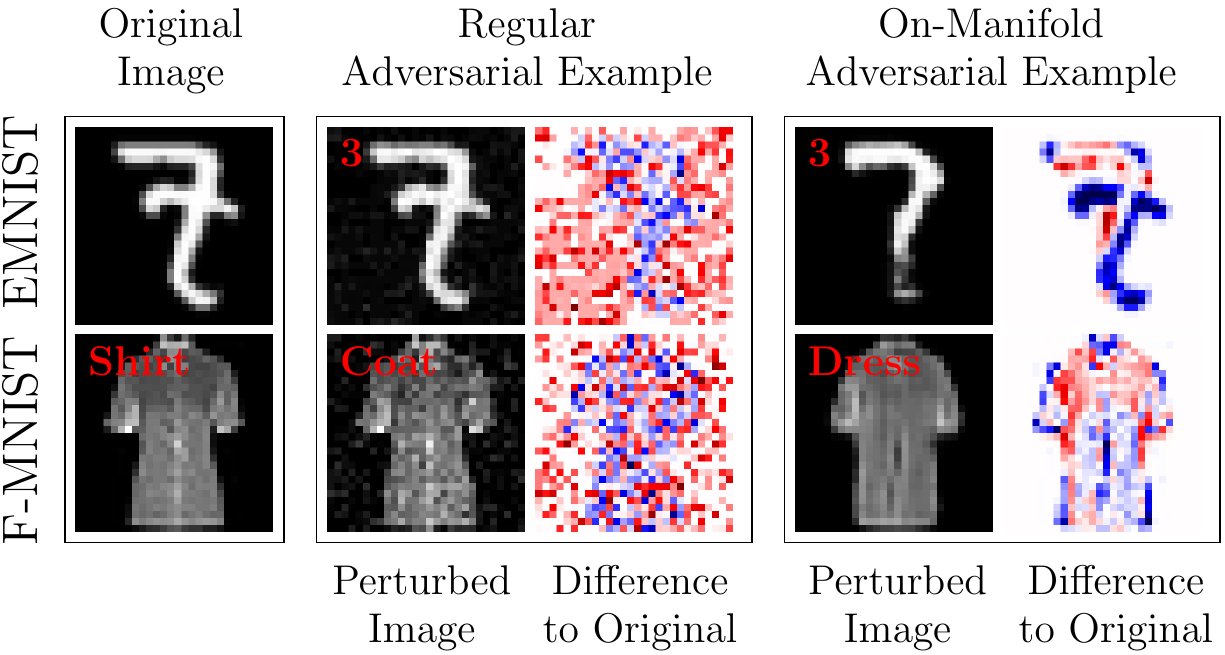}
    \end{subfigure}
    \vskip -6px
    \caption{Adversarial examples, and their (normalized) difference to the original image, in the context of the underlying manifold, \eg, class manifolds ``5'' and ``6'' on \MNIST \cite{CohenARXIV2017}, allow to study their relation to generalization. Regular adversarial examples are not constrained to the manifold, \cf (a), and often result in (seemingly) random noise patterns; in fact, we show that they leave the manifold. However, adversarial examples on the manifold can be found as well, \cf (b), resulting in meaningful manipulations of the image content; however, care needs to be taken that the actual, true label \wrt the manifold does not change, \cf (c).}
    \label{fig:introduction}
    \vskip -2px
\end{figure}

Adversarial robustness describes a deep network's ability to defend against adversarial examples \cite{SzegedyARXIV2013}, imperceptibly perturbed images causing mis-classification. These adversarial attacks pose severe security threats, as demonstrated against Clarifai.com~\cite{LiuICLR2017,BhagojiARXIV2017b} or Google Cloud Vision~\cite{IlyasICML2018}. Despite these serious risks, defenses against such attacks have been largely ineffective; only adversarial training, \ie, training on adversarial examples \cite{MadryICLR2018,GoodfellowARXIV2014}, has been shown to work well in practice \cite{AthalyeARXIV2018,AthalyeARXIV2018b} -- at the cost of computational overhead and reduced accuracy. Overall, the problem of adversarial robustness is left open and poorly understood -- even for simple datasets such as \MNIST \cite{CohenARXIV2017} and \FashionMNIST \cite{XiaoARXIV2017}.

The phenomenon of adversarial examples itself, \ie, their mere existence, has also received considerable attention. Recently, early explanations, \eg, attributing adversarial examples to ``rare pockets'' of the classification surface \cite{SzegedyARXIV2013} or linearities in deep networks \cite{GoodfellowARXIV2014}, have been superseded by the manifold assumption \cite{GilmerICLRWORK2018,TanayARXIV2016}: adversarial examples are assumed to leave the underlying, low-dimensional but usually unknown data manifold. \red{However, only \cite{SongICLR2018} provide experimental evidence supporting this assumption.} Yet, on a simplistic toy dataset, Gilmer \etal \cite{GilmerICLRWORK2018} also found adversarial examples on the manifold, as also tried on real datasets \cite{SongARXIV2018,BrownARXIV2018,ZhaoICLR2018}, rendering the manifold assumption questionable. Still, the manifold assumption fostered research on novel defenses \cite{IlyasARXIV2017,SamangoueiICLR2018,SchottARXIV2018}.

Beyond the existence of adversarial examples, their relation to generalization is an important open problem. Recently, it has been argued \cite{TsiprasARXIV2018,SuARXV2018} that there exists an inherent trade-off, \ie, robust and accurate models seem impossible. While Tsipras \etal \cite{TsiprasARXIV2018} provide a theoretical argument on a toy dataset, Su \etal \cite{SuARXV2018} evaluate the robustness of different models on ImageNet \cite{RussakovskyIJCV2015}. However, these findings have to be questioned  given the results in \cite{GilmerICLRWORK2018,RozsaICMLA2016} showing the opposite, \ie, better generalization helps robustness.

In order to address this controversy, and in contrast to \cite{TsiprasARXIV2018,SuARXV2017,RozsaICMLA2016}, we consider adversarial robustness in the context of the underlying manifold. In particular, to break the hypothesis down, we explicitly ask whether adversarial examples leave, or stay on, the manifold. On \MNIST, for example, considering the class manifolds for ``5'' and ``6'', as illustrated in \figref{fig:introduction}, adversarial examples are not guaranteed to lie on the manifold, \cf \figref{fig:introduction} (a). Adversarial examples can, however, also be constrained to the manifold, \cf \figref{fig:introduction} (b); in this case, it is important to ensure that the adversarial examples do not actually change their label, \ie, are more likely to be a ``6'' than a ``5'', as in \figref{fig:introduction} (c). For clarity, we refer to unconstrained adversarial examples, as illustrated in \figref{fig:introduction} (a), as \emph{regular adversarial examples}; in contrast to adversarial examples constrained to the manifold, so-called \emph{on-manifold adversarial examples}.

\myparagraph{Contributions:} Based on this distinction between regular robustness, \ie, against regular, unconstrained adversarial examples, and on-manifold robustness, \ie, against adversarial examples constrained to the manifold, we show:
\begin{enumerate}[itemsep=0ex,topsep=2px,parsep=2px]
    \item regular adversarial examples leave the manifold;
    \item adversarial examples constrained to the manifold, \ie, on-manifold adversarial examples, exist and can be computed using an approximation of the manifold;
    \item on-manifold robustness is essentially generalization;
    \item \red{and regular robustness and generalization are not necessarily contradicting goals, \ie, for any arbitrary but fixed model, better generalization through additional training data does not worsen robustness.}
\end{enumerate}
\noindent \red{We conclude that both robust and accurate models are possible and can, \eg, be obtained through adversarial training on larger training sets.} Additionally, we propose on-manifold adversarial training to boost generalization in settings where the manifold is known, can be approximated, or invariances of the data are known. We present experimental results on a novel MNIST-like, synthetic dataset with known manifold, as well as on \MNIST \cite{CohenARXIV2017}, \FashionMNIST \cite{XiaoARXIV2017} and \Celeb \cite{LiuICCV2015}. \red{We will make our code and data publicly available.}
\vskip 0px
\section{Related Work}
\label{sec:related-work}

\begin{figure*}
    \centering
    \vskip -0.4cm
    \includegraphics[width=1\textwidth]{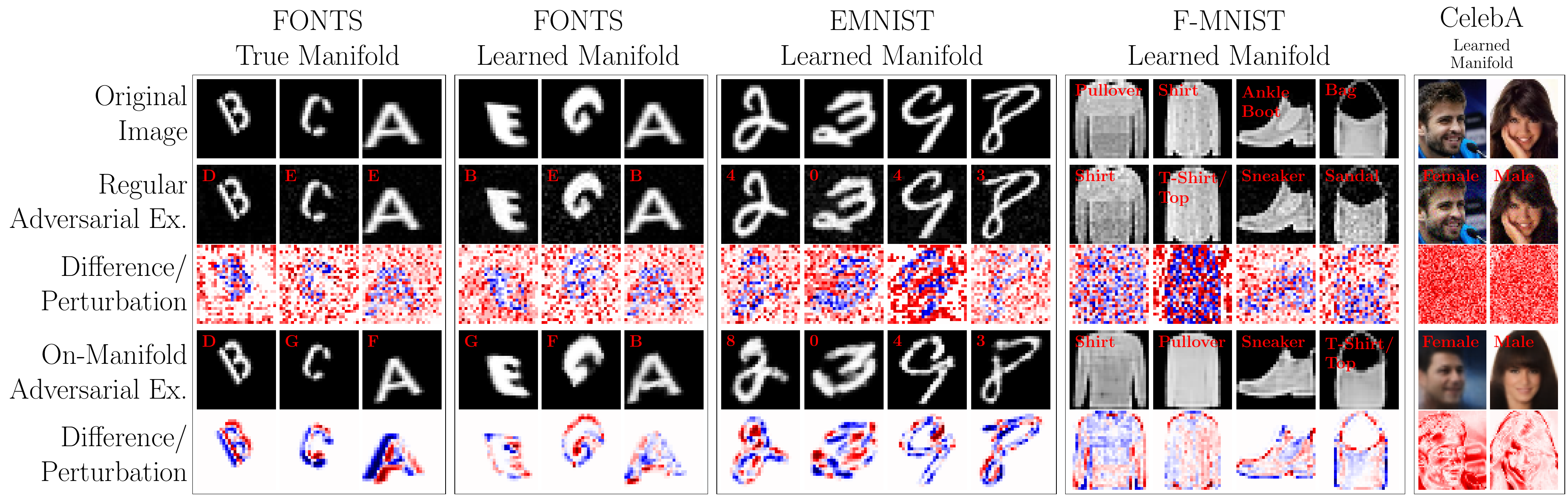}
    \vskip -6px 
    \caption{Regular and on-manifold adversarial examples on our synthetic dataset, \Fonts, consisting of randomly transformed characters ``A'' to ``J'', \MNIST \cite{CohenARXIV2017}, \Fashion \cite{XiaoARXIV2017} and \Celeb \cite{LiuICCV2015}. On \Fonts, the manifold is known by construction; in the other cases, the class manifolds have been approximated using \VAEGANs \cite{LarsenICML2016,RoscaARXIV2017}. The difference (normalized; or their magnitude on \Celeb) to the original test image reveals the (seemingly) random noise patterns of regular adversarial examples in contrast to reasonable concept changes of on-manifold adversarial examples.}
    \label{fig:main-examples}
\end{figure*}

\myparagraph{Attacks:} Adversarial examples for deep networks were first reported in \cite{SzegedyARXIV2013}; the problem of adversarial machine learning, however, has already been studied earlier~\cite{BiggioPR2018}. Adversarial attacks on deep networks range from white-box attacks \cite{SzegedyARXIV2013,GoodfellowARXIV2014,KurakinARXIV2016b,PapernotSP2016b,MoosaviCVPR2016,MadryICLR2018,CarliniSP2017,RoszaBMVC2017,DongCVPR2018,LuoAAAI2018}, with full access to the model (weights, gradients \etc), to black-box attacks \cite{ChenAISEC2017,BrendelARXIV2017a,SuARXV2017,IlyasARXIV2018,SarkarARXIV2017,NarodytskaCVPRWORK2017}, with limited access to model queries. White-box attacks based on first-order optimization, \eg, \cite{MadryICLR2018,CarliniSP2017}, are considered state-of-the-art. Due to their transferability \cite{LiuICLR2017,XieARXIV2018,PapernotASIACCS2017}, these attacks can also be used in a black-box setting (\eg using model stealing \cite{ShokriSP2017,PapernotASIACCS2017,TramerUSENIX2016,WangSP2018,OhICLR2018,JuutiARXIV2018}) and have, thus, become standard for evaluation. Recently, generative models have also been utilized to craft -- or learn -- more natural adversarial examples \cite{SongARXIV2018,BrownARXIV2018,ZhaoICLR2018,SchottARXIV2018}. Finally, adversarial examples have been applied to a wide variety of tasks, also beyond computer vision, \eg, \cite{FischerICLR2017,CisseNIPS2017,TabacofARXIV2016,KosSPWORK2018,HuangICLR2017,LinIJCAI2017,AlzantotEMNLP2018,CarliniWP2018}.

\myparagraph{Defenses:} Proposed defenses include detection and rejection methods \cite{GrosseARXIV2017,FeinmanARXIV2017,LiaoCVPR2018,MaARXIV2018,AmsalegWIFS2017,MetzenARXIV2017}, pre-processing, quantization and dimensionality reduction methods \cite{BuckmanICLR2018,PrakashDCC2018,BhagojiARXIV2017}, manifold-projection methods \cite{IlyasARXIV2017,SamangoueiICLR2018,SchottARXIV2018,ShenARXIV2017}, methods based on stochasticity/regularization or adapted architectures \cite{ZantedschiAISEC2017,BhagojiARXIV2017,NayebiARXIV2017,SimonGabrielARXIV2018,HeinNIPS2017,JakubovitzARXIV2018,RossAAAI2018,KannanARXIV2018,LambARXIV2018,XieICLR2018}, ensemble methods \cite{LiuARXIV2017,StraussARXIV2017,HeUSENIXWORK2017,TramerICLR2018}, as well as adversarial training \cite{ZantedschiAISEC2017,MiyatoICLR2016,HuangARXIV2015,ShahamNEUROCOMPUTING2018,SinhaICLR2018,LeeARXIV2017b,MadryICLR2018}; \red{however, many defenses have been broken, often by considering ``specialized'' or novel attacks \cite{CarliniAISec2017,CarliniARXIV2016,AthalyeARXIV2018b,AthalyeARXIV2018}}. In \cite{AthalyeARXIV2018}, only adversarial training, \eg, the work by Madry \etal \cite{MadryICLR2018}, has been shown to be effective -- although many recent defenses have not been studied extensively. Manifold-based methods, in particular, have received some attention lately: in \cite{IlyasARXIV2017,SamangoueiICLR2018}, generative adversarial networks \cite{GoodfellowNIPS2014} are used to project an adversarial example back to the learned manifold. Similarly, in \cite{SchottARXIV2018}, variational auto-encoders \cite{KingmaICLR2014} are used to perform robust classification.

\myparagraph{Generalization:} Research also includes independent benchmarks of attacks and defenses \cite{CarliniAISec2017,CarliniARXIV2016,AthalyeARXIV2018b,AthalyeARXIV2018,SharmaARXIV2017}, their properties \cite{LiuICLR2017,SharifARXIV2018}, as well as theoretical questions \cite{HeinNIPS2017,JakubovitzARXIV2018,FawziICMLWORK2015,TanayARXIV2016,GilmerICLRWORK2018,SimonGabrielARXIV2018,TsiprasARXIV2018,WangICML2018}. Among others, the existence of adversarial examples \cite{SzegedyARXIV2013,GoodfellowARXIV2014,TanayARXIV2016} raises many questions. While Szegedy \etal \cite{SzegedyARXIV2013} originally thought of adversarial examples as ``extremely'' rare negatives and Goodfellow \etal \cite{GoodfellowARXIV2014} attributed adversarial examples to the linearity in deep networks, others argued against these assumptions \cite{GilmerICLRWORK2018,TanayARXIV2016}. Instead, a widely accepted theory is the manifold assumption; adversarial examples are assumed to leave the data manifold \cite{GilmerICLRWORK2018,TanayARXIV2016,IlyasARXIV2017,SamangoueiICLR2018,SchottARXIV2018}.

This paper is particularly related to work on the connection of adversarial examples to generalization \cite{TsiprasARXIV2018,SuARXV2018,GilmerICLRWORK2018,RozsaICMLA2016}. Tsipras \etal \cite{TsiprasARXIV2018} and Su \etal \cite{SuARXV2018} argue that there exists an inherent trade-off between robustness and generalization. However, the theoretical argument in \cite{TsiprasARXIV2018} is questionable as adversarial examples are allowed to change their actual, true label \wrt the data distribution, as illustrated \figref{fig:introduction} (c). The experimental results obtained in \cite{SuARXV2018,RozsaICMLA2016} stem from comparing different architectures and training strategies; in contrast, we consider robustness and generalization for any arbitrary but fixed model. On a \red{simple} synthetic toy dataset, Gilmer \etal \cite{GilmerICLRWORK2018} show that on-manifold adversarial examples exist. We further show that on-manifold adversarial examples also exist on real datasets with unknown manifold, similar to \cite{ZhaoICLR2018}. \red{In contrast to \cite{GilmerICLRWORK2018,ZhaoICLR2018}, we utilize a gradient-based attack on the manifold, not in image space.} Our work is also related to \cite{FawziICIP2016} and \cite{MiyatoICLR2016,MiyatoPAMI2018} where variants of adversarial training are used to boost (semi-)supervised learning. While, \eg, Fawzi \etal \cite{FawziICIP2016}, apply adversarial training to image transformations, we further perform adversarial training on adversarial examples constrained to the true, or approximated, manifold. \red{This is also different from adversarial data augmentation schemes driven by GANs, \eg, \cite{RatnerNIPS2017,SixtFRAI2018,AntoniouICANN2018,CubukARXIV2018}, where training examples are generated, but without the goal to be mis-classified.} \red{Finally, \cite{SongICLR2018} provide experimental evidence that adversarial examples have low probability under the data distribution; we show that adversarial examples have, in fact, zero probability.}
\section{Disentangling Adversarial Robustness\\[2px]\hspace*{-6px}and Generalization}
\label{sec:main}

To clarify the relationship between adversarial robustness and generalization, we explicitly distinguish between regular and on-manifold adversarial examples, as illustrated in \figref{fig:introduction}. Then, the hypothesis \cite{TsiprasARXIV2018,SuARXV2018} that robustness and generalization are contradicting goals is challenged in four arguments: regular \red{unconstrained} adversarial examples leave the manifold; adversarial examples constrained to the manifold exist; robustness against on-manifold adversarial examples is essentially generalization; \red{robustness against regular adversarial examples is not influenced by generalization when controlled through the amount of training data}. Altogether, our results imply that adversarial robustness and generalization are not opposing objectives \red{and both robust and accurate models are possible} but require higher sample complexity.

\vskip 0px
\subsection{Experimental Setup}

\myparagraph{Datasets:} We use \MNIST \cite{CohenARXIV2017}, F(ashion)-MNIST \cite{XiaoARXIV2017} and \Celeb \cite{LiuICCV2015} for our experiments ($240\text{k}/40\text{k}$, $60\text{k}/10\text{k}$ and $182\text{k}/20\text{k}$ training/test images); \Celeb has been re-sized to $56{\times}48$ and we classify ``Male'' \vs ``Female''. Our synthetic dataset, \Fonts, consists of letters ``A'' to ``J'' of $1000$ Google Fonts randomly transformed (uniformly over translation, shear, scale, rotation in $[-0.2,0.2]$, $[-0.5,0.5]$, $[0.75,1.15]$, $[-\nicefrac{\pi}{2},\nicefrac{\pi}{2}]$) using a spatial transformer network \cite{JaderbergNIPS2015} such that the generation process is completely differentiable. The latent variables correspond to the transformation parameters, font and class. We generated $960\text{k}/40\text{k}$ (balanced) training/test images of size $28{\times}28$.

We consider classifiers with three (four on \Celeb) convolutional layers ($4\times4$ kernels; stride $2$; $16$, $32$, $64$ channels), each followed by ReLU activations and batch normalization \cite{IoffeICML2015}, and two fully connected layers. The networks are trained using ADAM \cite{KingmaICLR2015}, with learning rate $0.01$ (decayed by $0.95$ per epoch), weight decay $0.0001$ and batch size $100$, for $20$ epochs. Most importantly, to control their generalization performance, we use $N$ training images, with $N$ between $250$ and $40\text{k}$; for each $N$, we train $5$ models with random weight initialization \cite{GlorotAISTATS2010} an report averages.

We learn class-specific \VAEGANs, similar to \cite{LarsenICML2016,RoscaARXIV2017}, to approximate the underlying manifold; we refer to the supplementary material for details.

\myparagraph{Attack:} Given an image-label pair $(x,y)$ from an unknown data distribution $p$ and a classifier $f$, an adversarial example is a perturbed image $\tilde{x} = x + \delta$ which is mis-classified by the model, \ie, $f(\tilde{x}) \neq y$. While our results can be confirmed using other attacks and norms (see the supplementary material for \cite{CarliniSP2017} and transfer attacks), for clarity, we concentrate on the $L_{\infty}$ white-box attack by Madry \etal \cite{MadryICLR2018} that directly maximizes the training loss,
\vskip -14px
\begin{align}
	\max_\delta \cL(f(x + \delta), y)\quad\text{s.t.}\quad\|\delta\|_{\infty} \leq \epsilon, \tilde{x}_i \in [0,1],\label{eq:main-off-manifold-attack}
\end{align}
\vskip -2px
\noindent using projected gradient descent; where $\cL$ is the cross-entropy loss and $\tilde{x}=x+\delta$. The $\epsilon$-constraint is meant to ensure perceptual similarity. We run $40$ iterations of ADAM \cite{KingmaICLR2015} with learning rate $0.005$ and consider $5$ restarts, (distance and direction) uniformly sampled in the $\epsilon$-ball for $\epsilon = 0.3$. Optimization is stopped as soon as the predicted label changes, \ie, $f(\tilde{x}) \neq y$. We attack $1000$ test images.

\begin{figure}
    \centering
    \vskip -0.3cm
    \hskip -0.1cm
    \begin{subfigure}[t]{0.5\textwidth}
        \begin{subfigure}[t]{0.49\textwidth}
            \includegraphics[width=0.9\textwidth]{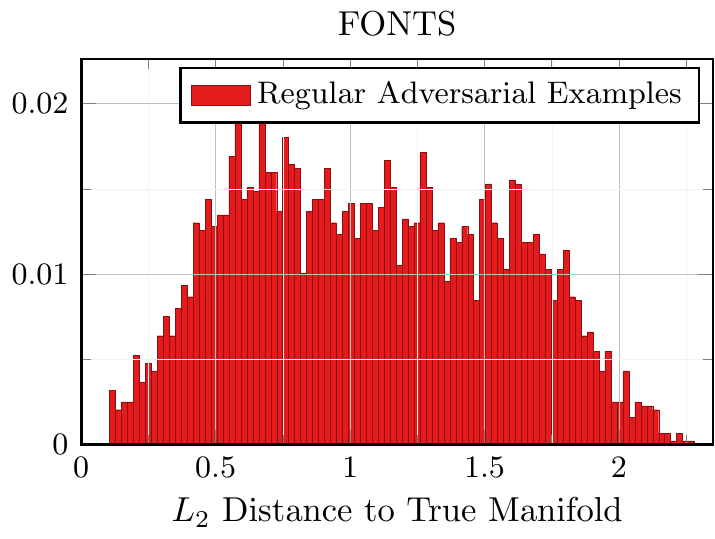}
        \end{subfigure}
        \begin{subfigure}[t]{0.49\textwidth}
            \includegraphics[width=0.9\textwidth]{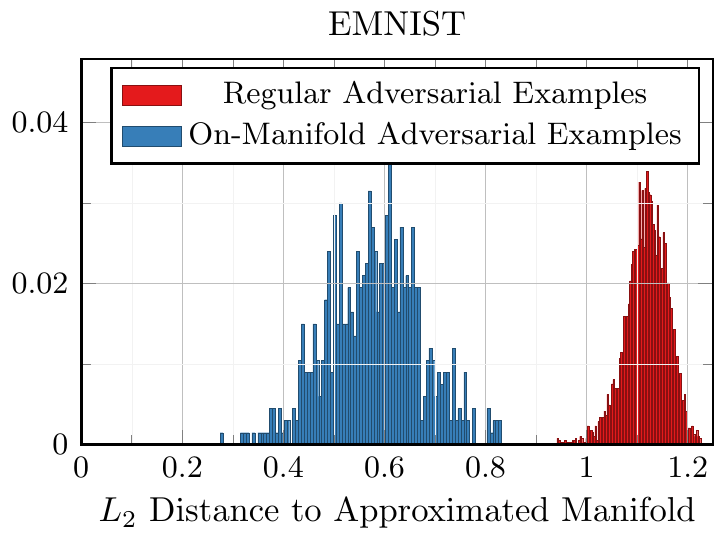}
        \end{subfigure}
    \end{subfigure}
    \vskip -6px
    \caption{Distance of adversarial examples to the true, on \Fonts (left), or approximated, on \MNIST (right), manifold. We show normalized histograms of the $L_2$ distance of adversarial examples to their projections onto the manifold ($4377$/$3837$ regular adversarial examples on \Fonts/\MNIST; $667$ on-manifold adversarial examples on \MNIST). Regular adversarial examples exhibit a significant distance to the manifold; on \MNIST, clearly distinguishable from on-manifold adversarial examples.}
    \label{fig:main-hypo1}
    \vskip 0px
\end{figure}

\myparagraph{Adversarial Training:} An established defense is adversarial training, \ie, training on adversarial examples crafted during training \cite{ZantedschiAISEC2017,MiyatoICLR2016,HuangARXIV2015,ShahamNEUROCOMPUTING2018,SinhaICLR2018,LeeARXIV2017b,MadryICLR2018}. Madry \etal \cite{MadryICLR2018} consider the min-max problem
\vskip -14px
\begin{align}
	\hskip -8px \min_w \sum_{n = 1}^N{\max_{\|\delta\|_{\infty} \leq \epsilon,x_{n,i}{+}\delta_i\in[0,1]}}{\cL}(f(x_n{+}\delta; w), y_n)
	\label{eq:main-off-manifold-adversarial-training}
\end{align}
\vskip -2px
\noindent where $w$ are the classifier's weights and $x_n$ the training images. As shown in the supplementary material, we considered different variants \cite{SzegedyARXIV2013,GoodfellowARXIV2014,MadryICLR2018}; in the paper, however, we follow common practice and train on $50\%$ clean images and $50\%$ adversarial examples \cite{SzegedyARXIV2013}. For $\epsilon = 0.3$, the attack (for the inner optimization problem) is run for full $40$ iterations, \ie, is not stopped at the first adversarial example found. Robustness of the obtained network is measured by computing the attack \textbf{success rate}, \ie, the fraction of successful attacks on correctly classified test images, as, \eg, in \cite{CarliniSP2017}, for a fixed~$\epsilon$; lower success rate indicates higher robustness of the network.

\vskip 0px
\subsection{Adversarial Examples Leave the Manifold}

The idea of adversarial examples leaving the manifold is intuitive on \MNIST where particular background pixels are known to be constant, see \figref{fig:main-examples}. If an adversarial example $\tilde{x}$ manipulates these pixels, it has zero probability under the data distribution and its distance to the manifold, \ie, the distance to its projection $\pi(\tilde{x})$ onto the manifold, should be non-zero. On \Fonts, with known generative process in the form of a decoder $\dec$ mapping latent variables $z$ to images $x$, the projection is obtained iteratively: $\pi(\tilde{x}) = \dec(\tilde{z})$ with $\tilde{z} = \argmin_{z} \|\dec(z) - \tilde{x})\|_2$ and $z$ constrained to valid transformations (font and class, known from the test image $x$, stay constant). On \MNIST, as illustrated in \red{\figref{fig:main-illustration-2} (right)}, the manifold is approximated using $50$ nearest neighbors; the projection $\pi(\tilde{x})$ onto the sub-space spanned by the $x$-centered nearest neighbors is computed through least squares. On both \Fonts and \MNIST, the distance $\|\tilde{x} - \pi(\tilde{x})\|_2$ is considered to asses whether the adversarial example $\tilde{x}$ actually left the manifold.

On \Fonts, \figref{fig:main-hypo1} (left) shows that regular adversarial examples clearly exhibit non-zero distance to the manifold. In fact, the projections of these adversarial examples to the manifold are almost always the original test images; as a result, the distance to the manifold is essentially the norm of the corresponding perturbation: $\|\tilde{x} - \pi(\tilde{x})\|_2 \approx \|\tilde{x} - x\|_2 = \|\delta\|_2$. This suggests that the adversarial examples leave the manifold in an almost orthogonal direction. On \MNIST, in \figref{fig:main-hypo1} (right), these results can be confirmed in spite of the crude local approximation of the manifold. Again, regular adversarial examples seem to leave the manifold almost orthogonally, \ie, their distance to the manifold coincides with the norm of the corresponding perturbations. These results show that regular adversarial examples essentially \emph{are} off-manifold adversarial examples; this finding is intuitive as for well-trained classifiers, leaving the manifold should be the ``easiest'' way to fool it; \red{results on \Fashion as well as a more formal statement of this intuition can be found in the supplementary material.}

\vskip 0px
\subsection{On-Manifold Adversarial Examples}

Given that regular adversarial examples leave the manifold, we intend to explicitly compute on-manifold adversarial examples. To this end, we assume our data distribution $p(x,y)$ to be conditional on the latent variables $z$, \ie, $p(x,y|z)$, corresponding to the underlying, low-dimensional manifold. \red{On this manifold, however, there is no notion of ``perceptual similarity'' in order to ensure label invariance, \ie, distinguish valid on-manifold adversarial examples, \figref{fig:introduction} (b), from invalid ones that change the actual, true label, \figref{fig:introduction} (c):}

\begin{definition}[On-Manifold Adversarial Example]
    Given the data distribution $p$, an on-manifold adversarial example for $x$ with label $y$ is a perturbed version $\tilde{x}$ such that $f(\tilde{x}) \neq y$ but $p(y | \tilde{x}) > p(y' | \tilde{x}) \forall y' \neq y$.\label{def:main-on-manifold-adversarial-example}
\end{definition}
\vskip 2px

\noindent Note that the posteriors $p(y|\tilde{x})$ correspond to the true, unknown data distribution; any on-manifold adversarial example $\tilde{x}$ violating \defref{def:main-on-manifold-adversarial-example} changed its actual, true label.

\begin{figure}[t]
    \centering
    \vskip -0.3cm
    \begin{subfigure}{0.28\textwidth}
        \includegraphics[width=1\textwidth]{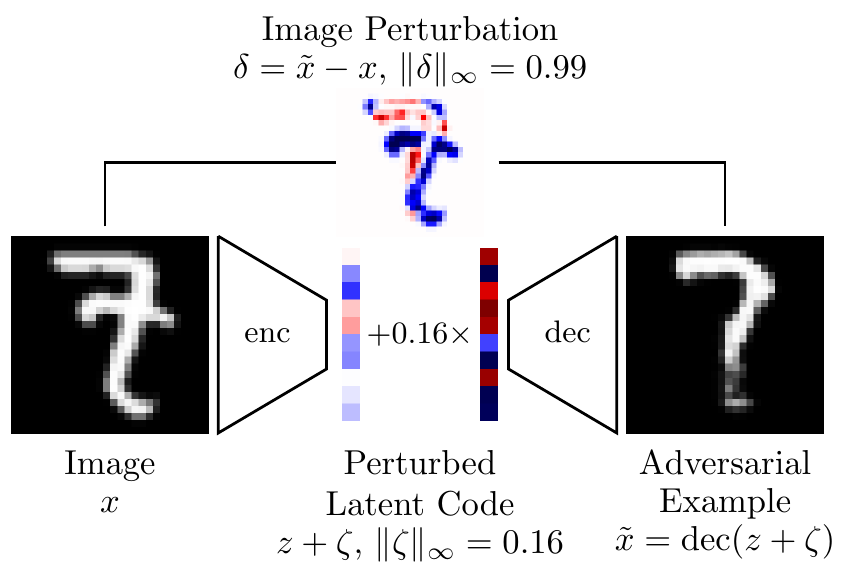}
    \end{subfigure}
    \begin{subfigure}{0.18\textwidth}
        \includegraphics[width=1\textwidth,trim={0 0 1.9cm 0},clip]{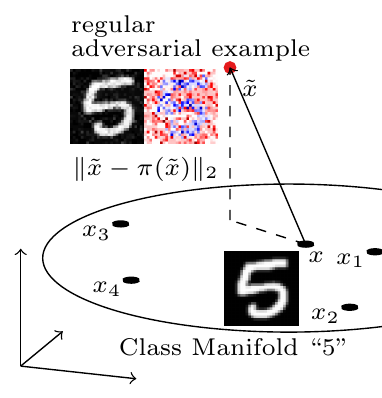}
    \end{subfigure}
    \vskip -8px
    \caption{\red{Left: On-manifold adversarial examples can be computed using learned, class-specific \VAEGANs \cite{LarsenICML2016,RoscaARXIV2017}. The perturbation $\zeta$ is obtained via \eqnref{eq:main-on-manifold-attack} and added to the latent code $z = \enc(x)$ yielding the adversarial example $\tilde{x} = \dec(z + \zeta)$ with difference $\delta = \tilde{x} - x$ in image space. Right: The distance of a regular adversarial example $\tilde{x}$ to the manifold, approximated using nearest neighbors, is computed as the distance to its orthogonal projection $\pi(\tilde{x})$: $\|\tilde{x} - \pi(\tilde{x})\|_2$. Large distances indicate that the adversarial example likely left the manifold.}}
    \label{fig:main-illustration-2}
    \vskip 0px
\end{figure}

In practice, we assume access to an encoder and decoder modeling the (class-conditional) distributions $p(z|x,y)$ and $p(x|z,y)$ -- in our case, achieved using \VAEGANs \cite{LarsenICML2016,RoscaARXIV2017}. Then, given the encoder \red{$\enc$ and decoder $\dec$} and as illustrated in \figref{fig:main-illustration-2} (left), we obtain the latent code $z = \enc(x)$ and compute the perturbation $\zeta$ by maximizing:
\vskip -14px
\begin{align}
\max_\zeta \cL(f(\dec(z + \zeta)), y)\quad\text{s.t.}\quad\|\zeta\|_{\infty}\leq \eta.\label{eq:main-on-manifold-attack}
\end{align}
\vskip -2px
\noindent The image-constraint, \ie, $\dec(z + \zeta) \in[0,1]$, is enforced by the decoder; the $\eta$-constraint can, again, be enforced by projection; and we can additionally enforce a constraint on $z + \zeta$, \eg, corresponding to a prior on $z$. Label invariance, as in \defref{def:main-on-manifold-adversarial-example}, is ensured by considering only class-specific encoders and decoders, \ie, the data distribution is approximated per class. We use $\eta = 0.3$ and the same optimization procedure as for \eqnref{eq:main-off-manifold-attack}; on approximated manifolds, the perturbation $z + \zeta$ is additionally constrained to $[-2,2]^{10}$, corresponding to a truncated normal prior from the class-specific \VAEGANs; we attack $2500$ test images.

On-manifold adversarial examples obtained through \eqnref{eq:main-on-manifold-attack} are similar to those crafted in \cite{GilmerICLRWORK2018}, \cite{SchottARXIV2018}, \cite{AthalyeARXIV2018} or \cite{ZhaoICLR2018}. However, in contrast to \cite{GilmerICLRWORK2018,SchottARXIV2018,AthalyeARXIV2018}, we directly compute the perturbation $\zeta$ on the manifold instead of computing the perturbation $\delta$ in the image space and subsequently projecting $x + \delta$ to the manifold. Also note that enforcing any similarity constraint through a norm on the manifold is significantly more meaningful compared to using a norm on the image space, as becomes apparent when comparing the obtained on-manifold adversarial examples in \figref{fig:main-examples} to their regular counterparts. Compared to \cite{ZhaoICLR2018}, we find on-manifold adversarial examples using a gradient-based approach instead of randomly sampling the latent space.

\figref{fig:main-examples} shows on-manifold adversarial examples for all datasets, which we found significantly harder to obtain compared to their regular counterparts. On \Fonts, using the true, known class manifolds, on-manifold adversarial examples clearly correspond to transformations of the original test image -- reflecting the true latent space. For the learned class manifolds, the perturbations are less pronounced, often manipulating boldness or details of the characters. Due to the approximate nature of the learned \VAEGANs, these adversarial examples are strictly speaking not always part of the true manifold -- as can be seen for the irregular ``A'' (\figref{fig:main-examples}, 6th column). On \MNIST and \Fashion, on-manifold adversarial examples represent meaningful manipulations, such as removing the tail of a hand-drawn ``8'' (\figref{fig:main-examples}, 10th column) or removing the collar of a pullover (\figref{fig:main-examples}, 11th column), in contrast to the random noise patterns of regular adversarial examples. However, these usually incur a smaller change in the images space; which also explains why regular, unconstrained adversarial examples almost always leave the manifold. Still, on-manifold adversarial examples are perceptually close to the original images. On \Celeb, the quality of on-manifold adversarial examples is clearly limited by the approximation quality of our \VAEGANs. Finally, \figref{fig:main-hypo1} (right) shows that on-manifold adversarial examples are closer to the manifold than regular adversarial examples -- in spite of the crude approximation of the manifold on \MNIST.

\vskip 0px
\subsection{On-Manifold Robustness is Essentially\\[2px]\hspace*{-5px}Generalization}

\begin{figure*}
    \centering
    \vskip -0.3cm
    \hskip -0.25cm
    \begin{subfigure}[t]{0.24\textwidth}
        \vskip 0px
        \centering
        \includegraphics[width=1\textwidth]{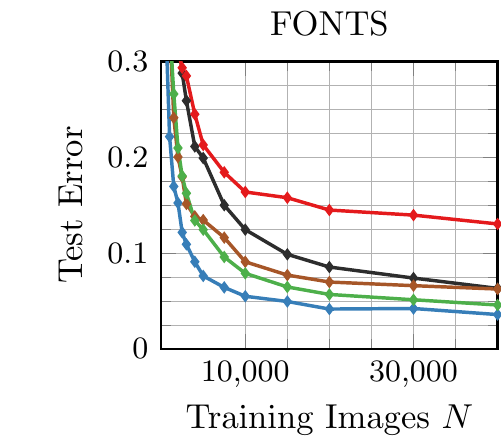}
    \end{subfigure}
    \begin{subfigure}[t]{0.24\textwidth}
        \vskip 0px
        \centering
        \includegraphics[width=1\textwidth]{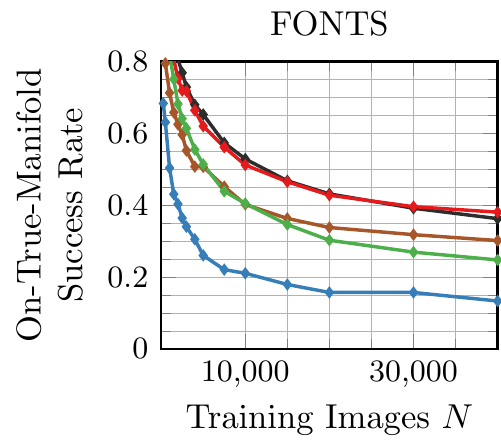}
    \end{subfigure}
    \begin{subfigure}[t]{0.24\textwidth}
        \vskip 0px
        \centering
        \includegraphics[width=1\textwidth]{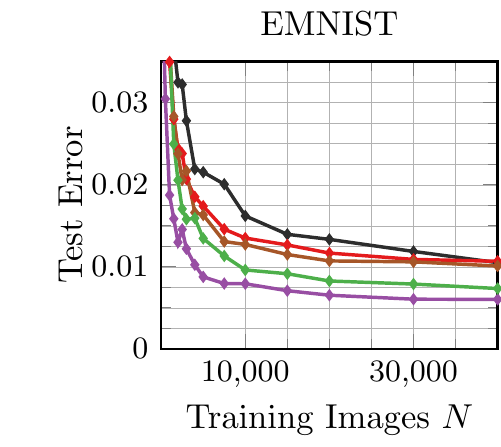}
    \end{subfigure}
    \begin{subfigure}[t]{0.24\textwidth}
        \vskip 0px
        \centering
        \includegraphics[width=1\textwidth]{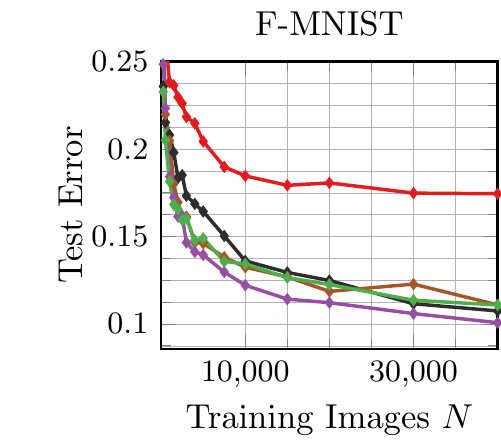}
    \end{subfigure}
    \\
    \hskip -0.25cm
    \begin{subfigure}[t]{0.24\textwidth}
        \vskip 0px
        \centering
        \includegraphics[width=1\textwidth]{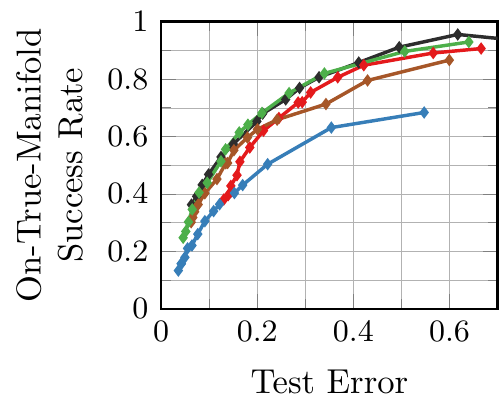}
    \end{subfigure}
    \begin{subfigure}[t]{0.24\textwidth}
        \vskip 0px
        \centering
        \includegraphics[width=1\textwidth]{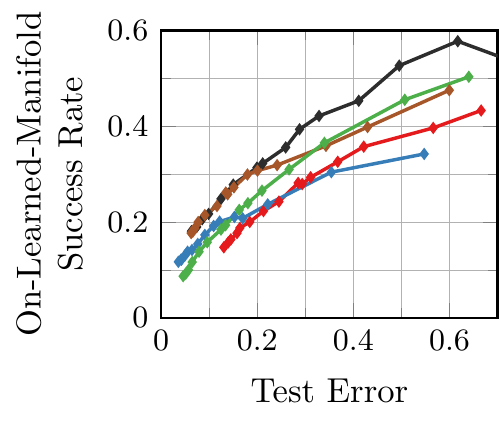}
    \end{subfigure}
    \begin{subfigure}[t]{0.24\textwidth}
        \vskip 0px
        \centering
        \includegraphics[width=1\textwidth]{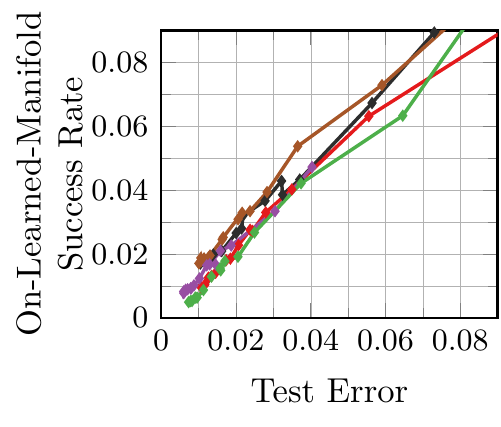}
    \end{subfigure}
    \begin{subfigure}[t]{0.24\textwidth}
        \vskip 0px
        \centering
        \includegraphics[width=1\textwidth]{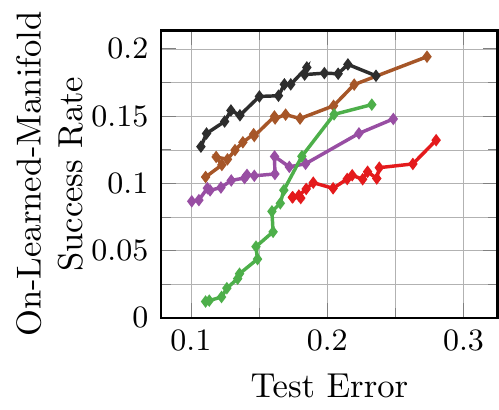}
    \end{subfigure}
    \\
    \fcolorbox{black!50}{white}{
    \begin{subfigure}[t]{0.975\textwidth}
        \centering
        \includegraphics[width=0.9\textwidth]{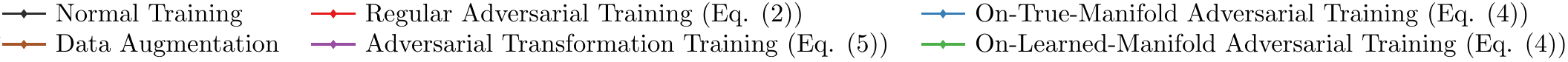}
    \end{subfigure}
    }
    \vskip -6px
    \caption{On-manifold robustness is strongly related to generalization, as shown on \Fonts, \MNIST and \Fashion considering on-manifold success rate and test error. Top: test error and on-manifold success rate in relation to the number of training images. As test error reduces, so does on-manifold success rate. Bottom: on-manifold success rate plotted against test error reveals the strong relationship between on-manifold robustness and generalization.}
    \label{fig:main-hypo3}
    \vskip 0px
\end{figure*}

We argue that on-manifold robustness is nothing different than generalization: as on-manifold adversarial examples have non-zero probability under the data distribution, they are merely generalization errors. This is shown in \figref{fig:main-hypo3} (top left) where test error and on-manifold success rate on \Fonts are shown. As expected, better generalization, \ie, using more training images $N$, also reduces on-manifold success rate. In order to make this relationship explicit, \figref{fig:main-hypo3} (bottom) plots on-manifold success rate against test error. Then, especially for \Fonts and \MNIST, the relationship of on-manifold robustness and generalization becomes apparent. On \Fashion, the relationship is less pronounced because on-manifold adversarial examples, computed using our \VAEGANs, are not close enough to real generalization errors. However, even on \Fashion, the experiments show a clear relationship between on-manifold robustness and generalization.

\vskip 0px
\subsubsection{On-Manifold Adversarial Training\\[2px]Boosts Generalization}

Given that generalization positively influences on-manifold robustness, we propose to adapt adversarial training to the on-manifold case in order to boost generalization:
\vskip -14px
\begin{align}
    \min_w \sum_{n=1}^N \max_{\|\zeta\|_{\infty} \leq \eta} \cL(f(\dec(z_n + \zeta); w), y_n).
    \label{eq:main-on-manifold-adversarial-training}
\end{align}
\vskip -2px
\noindent with $z_n = \dec(x_n)$ being the latent codes corresponding to training images $x_n$. Then, on-manifold adversarial training corresponds to robust optimization \wrt the true, or approximated, data distribution. For example, with \matthias{the} perfect decoder on \Fonts, the inner optimization problem finds ``hard'' images irrespective of their likelihood under the data distribution. For approximate $\dec$, the benefit of on-manifold adversarial training depends on how well the true data distribution is matched, \ie, how realistic the obtained on-manifold adversarial examples are; in our case, this depends on the quality of the learned \VAEGANs.

\begin{figure*}
    \centering
    \vskip -0.3cm
    \hskip -0.25cm
    \begin{subfigure}[t]{0.24\textwidth}
        \vskip 0px
        \centering
        \includegraphics[width=1\textwidth]{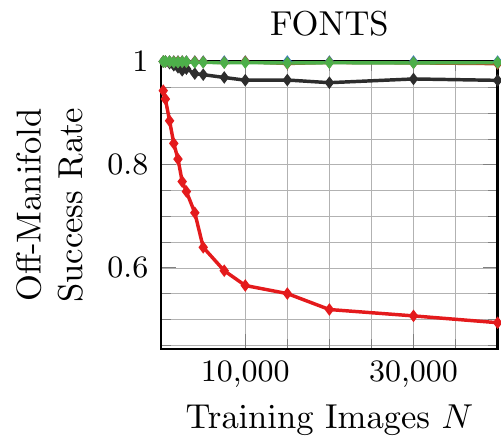}
    \end{subfigure}
    \begin{subfigure}[t]{0.24\textwidth}
        \vskip 0px
        \centering
        \includegraphics[width=1\textwidth]{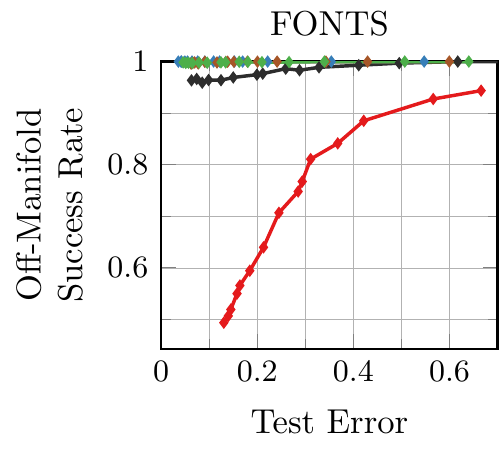}
    \end{subfigure}
    \begin{subfigure}[t]{0.24\textwidth}
        \vskip 0px
        \centering
        \includegraphics[width=1\textwidth]{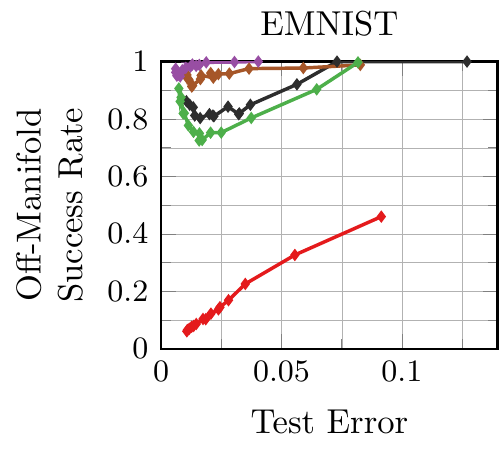}
    \end{subfigure}
    \begin{subfigure}[t]{0.24\textwidth}
        \vskip 0px
        \centering
        \includegraphics[width=1\textwidth]{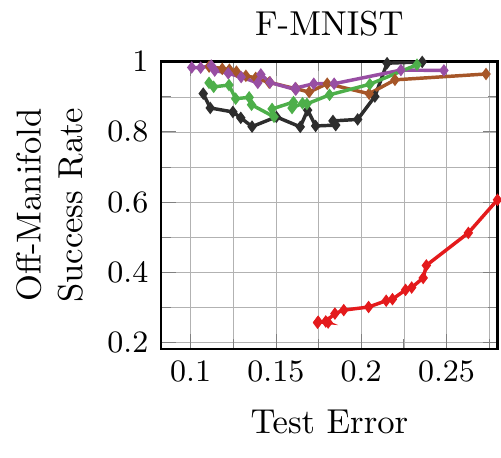}
    \end{subfigure}
    \\
    \fcolorbox{black!50}{white}{
    \begin{subfigure}[t]{0.975\textwidth}
        \centering
        \includegraphics[width=0.9\textwidth]{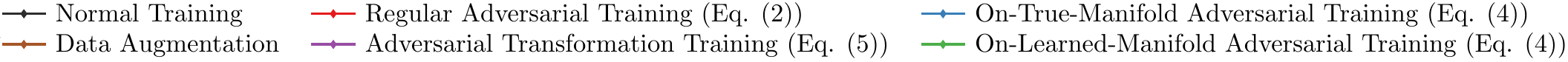}
    \end{subfigure}
    }
    \vskip -6px
    \caption{Regular robustness is not related to generalization, as demonstrated on \Fonts, \MNIST and \Fashion considering test error and (regular) success rate. On \Fonts (left), success rate is not influenced by test error, except for adversarial training. Plotting success rate against test error highlights the independence of robustness and generalization; however, different training strategies exhibit different robustness-generalization characteristics.}
    \label{fig:main-hypo4}
    \vskip 0px
\end{figure*}

Instead of approximating the manifold using generative models, we can exploit known invariances of the data. Then, adversarial training can be applied to these invariances, assuming that they are part of the true manifold. In practice, this can, for example, be accomplished using adversarial deformations \cite{AlaifariARXIV2018,XiaoICLR2018,EngstromARXIV2017}, \ie, adversarially crafted transformations of the image. For example, as on \Fonts, we consider $6$-degrees-of-freedom transformations corresponding to translation, shear, scaling and rotation:
\vskip -14px
\begin{align}
    \min_w \sum_{n = 1}^N \max_{\|t\|_{\infty} \leq \eta, t \in \mR^6} \cL(f(T(x_n; t); w), y_n).
    \label{eq:main-stn-adversarial-training}
\end{align}
\vskip -2px
\noindent where $T(x; t)$ denotes the transformation of image $x$ with parameters $t$ and the $\eta$-constraint ensures similarity and label invariance. Again, the transformations can be applied using spatial transformer networks \cite{JaderbergNIPS2015} such that $T$ is differentiable; $t$ can additionally be constrained to a reasonable space of transformations. We note that a similar approach has been used by Fawzi \etal \cite{FawziICIP2016} to boost generalization on, \eg, MNIST \cite{LecunIEEE1998}. However, the approach was considered as an adversarial variant of data augmentation and not motivated through the lens of on-manifold robustness. We refer to \eqnref{eq:main-stn-adversarial-training} as adversarial transformation training and note that, on \Fonts, this approach is equivalent to on-manifold adversarial training as the transformations coincide with the actual, true manifold by construction. We also include a data augmentation baseline, where the transformations $t$ are applied randomly.

We demonstrate the effectiveness of on-manifold adversarial training in \figref{fig:main-hypo3} (top). On \Fonts, with access to the true manifold, on-manifold adversarial training is able to boost generalization significantly, especially for low $N$, \ie, few training images. Our \VAEGAN approximation on \Fonts seems to be good enough to preserve the benefit of on-manifold adversarial training. On \MNIST and \Fashion, the benefit reduces with the difficulty of approximating the manifold; this is the ``cost'' of imperfect approximation. While the benefit is still significant on \MNIST, it diminishes on \Fashion. However, both on \MNIST and \Fashion, identifying invariances and utilizing adversarial transformation training recovers the boost in generalization; especially in contrast to the random data augmentation baseline. Overall, on-manifold adversarial training is a promising tool for improving generalization and we expect its benefit to increase with better generative models.

\vskip 0px
\subsection{Regular Robustness is Independent of\\[2px]\hspace*{-5px}Generalization}

\red{We argue that generalization, as measured \emph{on} the manifold \wrt the data distribution, is mostly independent of robustness against regular, possibly off-manifold, adversarial examples when varying the amount of training data}. Specifically, in \figref{fig:main-hypo4} (left) for \Fonts, it can be observed that -- except for adversarial training -- the success rate is invariant to the test error. \red{This can best be seen when plotting the success rate against test error for different numbers of training examples, \cf \figref{fig:main-hypo4} (middle left): only for adversarial training there exists a clear relationship; for the remaining training schemes success rate is barely influenced by the test error. In particular, better generalization does not worsen robustness.} Similar behavior can be observed on \MNIST and \Fashion, see \figref{fig:main-hypo4} (right). Here, it can also be seen that different training strategies exhibit different characteristics \wrt robustness and generalization. \red{Overall, regular robustness and generalization are not necessarily contradicting goals.}

As mentioned in \secref{sec:introduction}, these findings are in contrast to related work \cite{TsiprasARXIV2018,SuARXV2018} claiming that an inherent trade-off between robustness and generalization exists. For example, Tsipras \etal \cite{TsiprasARXIV2018} use a synthetic toy dataset to theoretically show that no model can be both robust and accurate (on this dataset). However, they allow the adversary to produce perturbations that change the actual, true label \wrt the data distribution, \ie, the considered adversarial examples are not adversarial examples according to \defref{def:main-on-manifold-adversarial-example}. Thus, it is unclear whether the suggested trade-off actually exists \red{for real datasets}; our experiments, \red{at least, as well as further analysis in the supplementary material} seem to indicate the contrary. Similarly, Su \etal \cite{SuARXV2018} experimentally show a trade-off between adversarial robustness and generalization by studying different models on ImageNet \cite{RussakovskyIJCV2015}. However, Su \etal compare the robustness and generalization characteristics of different models (\ie, different architectures, training strategies \etc), while we found that the generalization performance does not influence robustness for any \emph{arbitrary, but fixed} model.

\vskip 0px
\subsection{Discussion}
\label{subsec:main-discussion}

\begin{figure}
    \centering
    \vskip -0.3cm
    \hskip -0.25cm
    \begin{subfigure}[t]{0.235\textwidth}
        \vskip 0px
        \centering
        \includegraphics[width=1\textwidth]{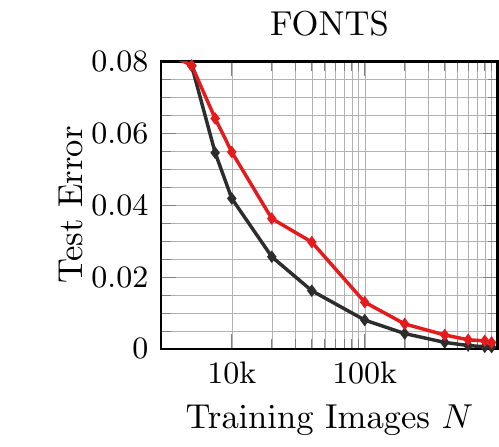}
    \end{subfigure}
    \begin{subfigure}[t]{0.235\textwidth}
        \vskip 0px
        \centering
        \includegraphics[width=1\textwidth]{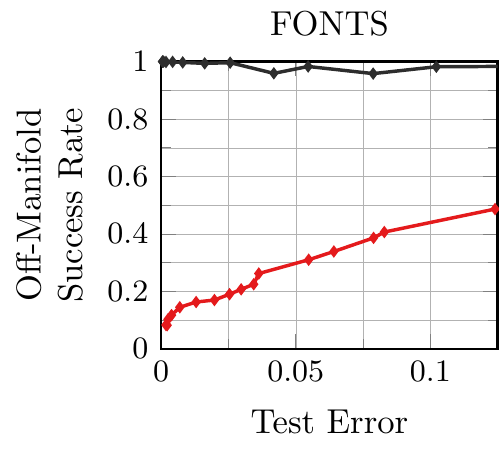}
    \end{subfigure}
    \\
    \begin{subfigure}[t]{0.235\textwidth}
        \vskip 0px
        \centering
        \includegraphics[width=1\textwidth]{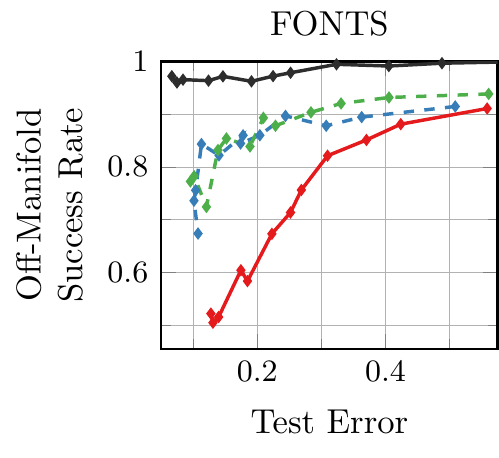}
    \end{subfigure}
    \begin{subfigure}[t]{0.235\textwidth}
        \vskip 0px
        \centering
        \includegraphics[width=1\textwidth]{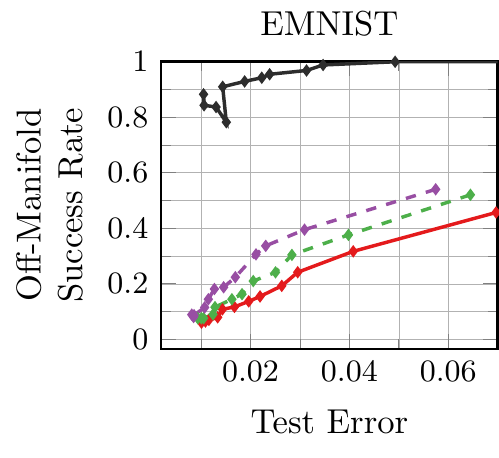}
    \end{subfigure}
    \\[-2px]
    \fcolorbox{black!50}{white}{
    \begin{subfigure}[t]{0.45\textwidth}
        \centering
        \includegraphics[width=1.025\textwidth]{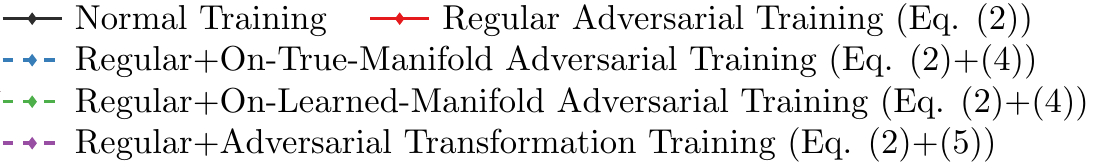}
    \end{subfigure}
    }
    \vskip -6px
    \caption{\red{Adversarial training on regular adversarial examples, potentially leaving the manifold, renders the learning problem more difficult. Top: With roughly $1.5$ to $2$ times the training data, adversarial training can still reach the same accuracy as normal training; results for ResNet-13~\cite{HeCVPR2016}. Bottom: Additionally, the trade-off can be controlled by combining regular and on-manifold adversarial training; results averaged over $3$ models.}}
    \label{fig:main-disc-2}
    \vskip 0px
\end{figure}

Our results imply that robustness and generalization are not \red{necessarily} conflicting goals, as believed in related work \cite{TsiprasARXIV2018,SuARXV2018}. This means, in practice, for any arbitrary but fixed model, better generalization will not worsen regular robustness. Different models (architectures, training strategies \etc) might, however, exhibit different robustness and generalization characteristics, as also shown in \cite{SuARXV2018,RozsaICMLA2016}. For adversarial training, on regular adversarial examples, the commonly observed trade-off between robustness and generalization is explained by the tendency of adversarial examples to leave the manifold. As result, the network has to learn (seemingly) random, but adversarial, noise patterns \emph{in addition} to the actual task at hand; rendering the learning problem harder. On simple datasets, such as \MNIST, these adversarial directions might avoid overfitting; on harder tasks, \eg, \Fonts or \Fashion, the discrepancy in test error between normal and adversarial training increases. \red{Our results also support the hypothesis that regular adversarial training has higher sample complexity \cite{SchmidtARXIV2018,KhouryARXIV2018}. In fact, on \Fonts, adversarial training can reach the same accuracy as normal training with roughly twice the amount of training data, as demonstrated in \figref{fig:main-disc-2} (top). Furthermore, as illustrated in \figref{fig:main-disc-2} (bottom), the trade-off between regular robustness and generalization can be controlled by combining regular and on-manifold adversarial training, \ie boost generalization while reducing robustness.}

\begin{figure}
    \centering
    \vskip -0.3cm
    \hskip -0.25cm
    \begin{subfigure}[t]{0.235\textwidth}
        \vskip 0px
        \centering
        \includegraphics[width=1\textwidth]{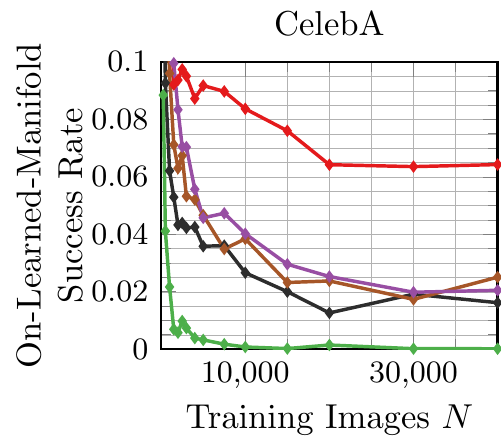}
    \end{subfigure}
    \begin{subfigure}[t]{0.235\textwidth}
        \vskip 0px
        \centering
        \includegraphics[width=1\textwidth]{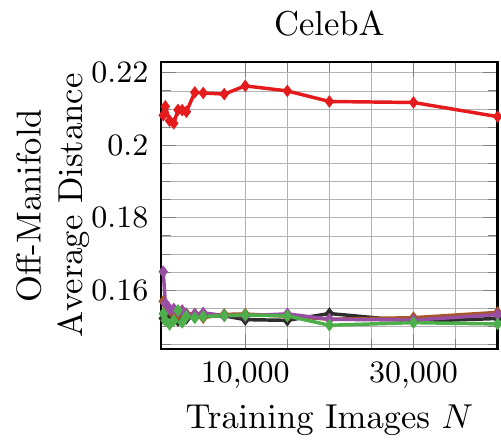}
    \end{subfigure}
    \\
    \hskip -0.25cm
    \begin{subfigure}[t]{0.235\textwidth}
        \vskip 0px
        \centering
        \includegraphics[width=1\textwidth]{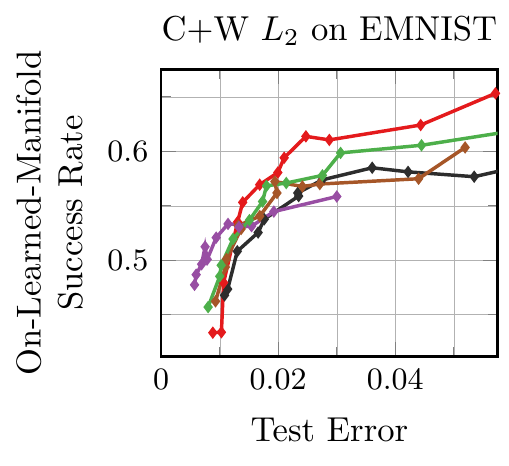}
    \end{subfigure}
    \begin{subfigure}[t]{0.235\textwidth}
        \vskip 0px
        \centering
        \includegraphics[width=1\textwidth]{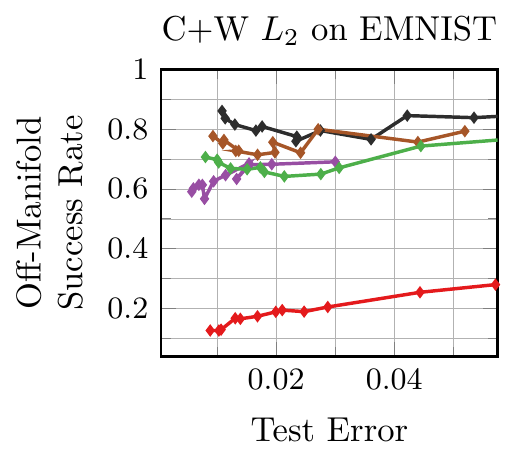}
    \end{subfigure}
    \\[-2px]
    \fcolorbox{black!50}{white}{
    \begin{subfigure}[t]{0.45\textwidth}
        \centering
        \includegraphics[width=1.01\textwidth]{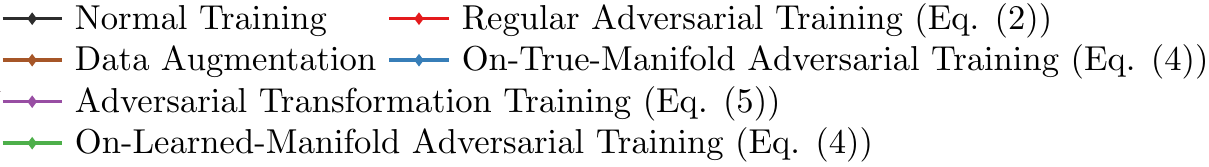}
    \end{subfigure}
    }
    \vskip -6px
    \caption{Results on \Celeb and using the $L_2$ Carlini and Wagner \cite{CarliniSP2017} attack. On \Celeb, as the class manifolds are significantly harder to approximate, the benefit of on-manifold adversarial training diminishes. For \cite{CarliniSP2017}, we used $120$ iterations; our hypotheses are confirmed, although \cite{CarliniSP2017} does not use the training loss as attack objective and the $L_2$ norm changes the similarity-constraint for regular and on-manifold adversarial examples.}
    \label{fig:main-disc-1}
    \vskip 0px
\end{figure}

The presented results can also be confirmed on more complex datasets, such as \Celeb, and using different threat models, \ie, attacks. On \Celeb, where \VAEGANs have difficulties approximating the manifold, \figref{fig:main-disc-1} (top left) shows that on-manifold robustness still improves with generalization although most on-manifold adversarial examples are not very realistic, see \figref{fig:main-examples}. Similarly, regular robustness, see \figref{fig:main-disc-1} (top right), is not influenced by generalization; here, we also show that the average distance of the perturbation, \ie, average $\|\delta\|_{\infty}$, when used to asses robustness leads to the same conclusions. Similarly, as shown in \figref{fig:main-disc-1} (bottom), our findings are confirmed using Carlini and Wagner's attack \cite{CarliniSP2017} with $L_2$-norm -- to show that the results generalize across norms. However, overall, we observed lower success rates using \cite{CarliniSP2017} and the $L_2$ norm. \red{Finally, our results can also be reproduced using transfer attacks (\ie, black-box attacks, which are generally assumed to be subsumed by white-box attacks \cite{AthalyeARXIV2018}) as well as and different architectures such as multi-layer perceptrons, ResNets~\cite{HeCVPR2016} and VGG~\cite{SimonyanARXIV2014}, as detailed in the supplementary material.}

\section{Conclusion}
\label{sec:conclusion}

In this paper, we intended to disentangle the relationship between adversarial robustness and generalization by initially adopting the hypothesis that robustness and generalization are contradictory \cite{TsiprasARXIV2018,SuARXV2018}. By considering adversarial examples in the context of the low-dimensional, underlying data manifold, we formulated and experimentally confirmed four assumptions. First, we showed that regular adversarial examples indeed leave the manifold, as widely assumed in related work \cite{GilmerICLRWORK2018,TanayARXIV2016,IlyasARXIV2017,SamangoueiICLR2018,SchottARXIV2018}. Second, we demonstrated that adversarial examples can also be found on the manifold, so-called on-manifold adversarial examples; even if the manifold has to be approximated, \eg, using \VAEGANs \cite{LarsenICML2016,RoscaARXIV2017}. Third, we established that robustness against on-manifold adversarial examples is clearly related to generalization. Our proposed on-manifold adversarial training exploits this relationship to boost generalization using an approximate manifold, or known invariances. Fourth, we provided evidence that \red{robustness against regular, unconstrained adversarial examples and generalization are not necessarily contradicting goals}: for any arbitrary but fixed model, better generalization, \eg, through more training data, does not reduce robustness.

{\small
\bibliographystyle{ieee_fullname}
\bibliography{bibliography}
}

\clearpage
\begin{appendix}
\twocolumn[{%
    \renewcommand\twocolumn[1][]{#1}%
    \begin{center}
        \centering
        \vskip -0.4cm
        \includegraphics[width=\textwidth]{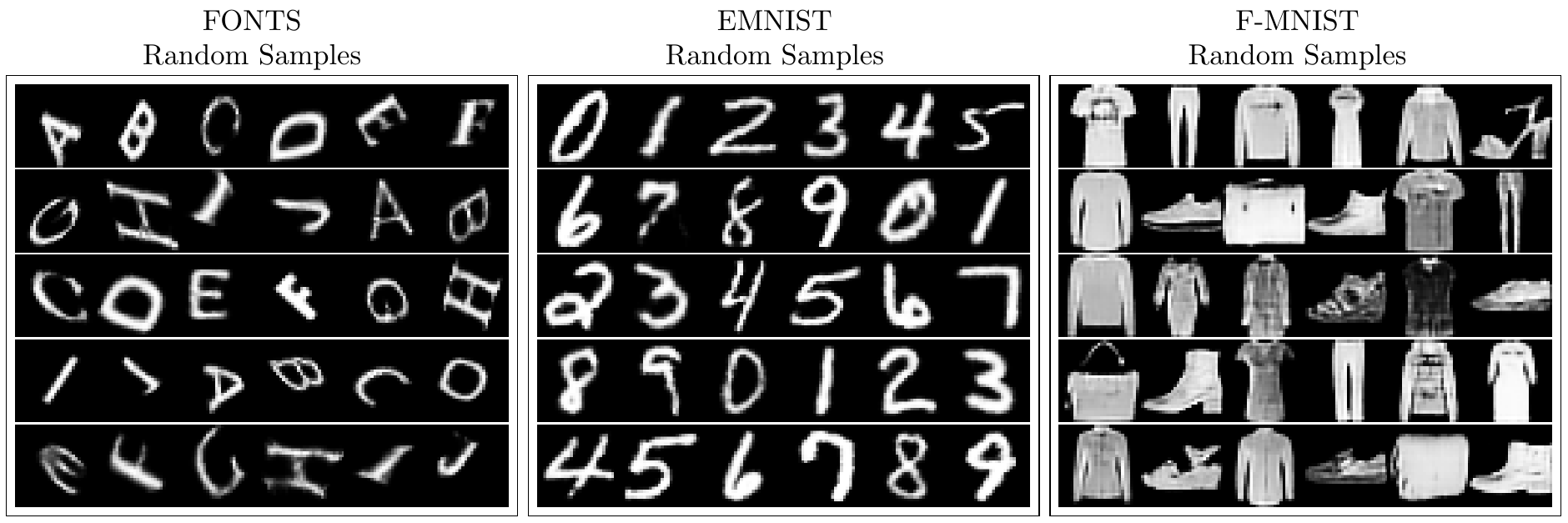}
        \vskip -6px
        \captionof{figure}{
            For \Fonts (left), \MNIST (middle) and \Fashion (right), we show random samples from the learned, class-specific \VAEGANs used to craft on-manifold adversarial examples. Our \VAEGANs generate realistic looking samples; although we also include problematic samples illustrating the discrepancy between true and approximated data distribution.
        }
        \label{fig:appendix-vaegan}
        \vskip 24px
    \end{center}
}]

\section{Overview}

In the main paper, we study the relationship between adversarial robustness and generalization. Based on the distinction between regular and on-manifold adversarial examples, we show that
\begin{enumerate*}
	\item regular adversarial examples leave the underlying manifold of the data;
	\item on-manifold adversarial examples exist;
	\item on-manifold robustness is essentially generalization;
	\item and regular robustness is independent of generalization.
\end{enumerate*}
For clarity and brevity, the main paper focuses on the $L_{\infty}$ attack by Madry \etal \cite{MadryICLR2018} and the corresponding adversarial training variant applied to simple
convolutional neural networks. For on-manifold adversarial examples, we approximate the manifold using class-specific \VAEGANs \cite{LarsenICML2016,RoscaARXIV2017}. In this document, we present comprehensive experiments demonstrating that our findings generalize across attacks, adversarial training variants, network architectures and to class-agnostic \VAEGANs.

\subsection{Contents}

In \secref{sec:appendix-setup}, we present additional details regarding our experimental setup, corresponding to Section 3.1 of the main paper: in \secref{sec:appendix-fonts}, we discuss details of our synthetic \Fonts datasets and, in \secref{sec:appendix-vaegan}, we discuss our \VAEGAN implementation. Then, in \secref{sec:appendix-off-manifold} we extend the discussion of Section 3.2 with further results demonstrating that adversarial examples leave the manifold. Subsequently, in \secref{sec:appendix-on-manifold}, we show and discuss additional on-manifold adversarial examples to supplement the examples shown in Fig.\ 2 of the main paper. Then, complementing the discussion in Sections 3.4 and 3.5, we consider additional attacks, network architectures and class-agnostic \VAEGANs. Specifically, in \secref{sec:appendix-attacks}, we consider the $L_2$ variant of the white-box attack by Madry \etal \cite{MadryICLR2018}, the $L_2$ white-box attack by Carlini and Wagner \cite{CarliniSP2017}, and black-box transfer attacks. In \secref{sec:appendix-network}, we present experiments on multi-layer perceptrons and, in \secref{sec:appendix-data-manifold}, we consider approximating the manifold using class-agnostic \VAEGANs. In \secref{sec:appendix-adversarian-training}, corresponding to Section 3.6, we consider different variants of regular and on-manifold adversarial training. Finally, in \secref{sec:appendix-adversarial-example}, we discuss our definition of adversarial examples in the context of related work by Tsipras \etal \cite{TsiprasARXIV2018}, as outlined in Section 3.5.

\begin{figure*}[t]
    \centering
    \vskip -0.4cm
    \begin{subfigure}[t]{0.32\textwidth}
        \centering
        \includegraphics[width=1\textwidth]{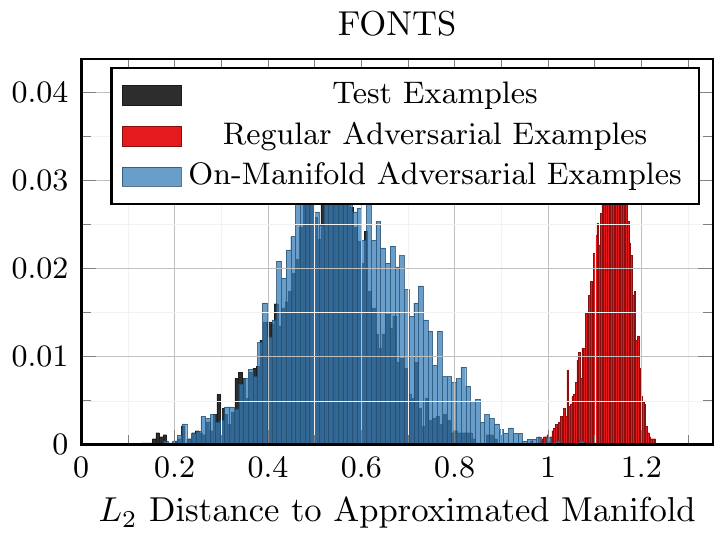}
    \end{subfigure}
    \begin{subfigure}[t]{0.32\textwidth}
        \centering
        \includegraphics[width=1\textwidth]{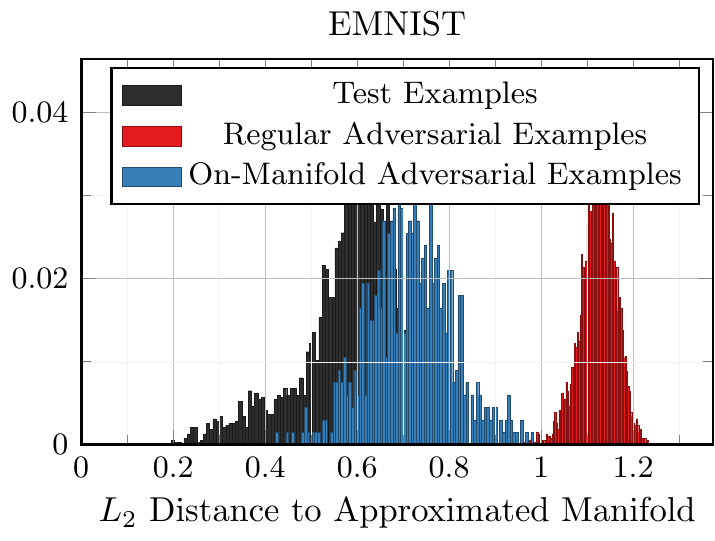}
    \end{subfigure}
    \begin{subfigure}[t]{0.32\textwidth}
        \centering
        \includegraphics[width=1\textwidth]{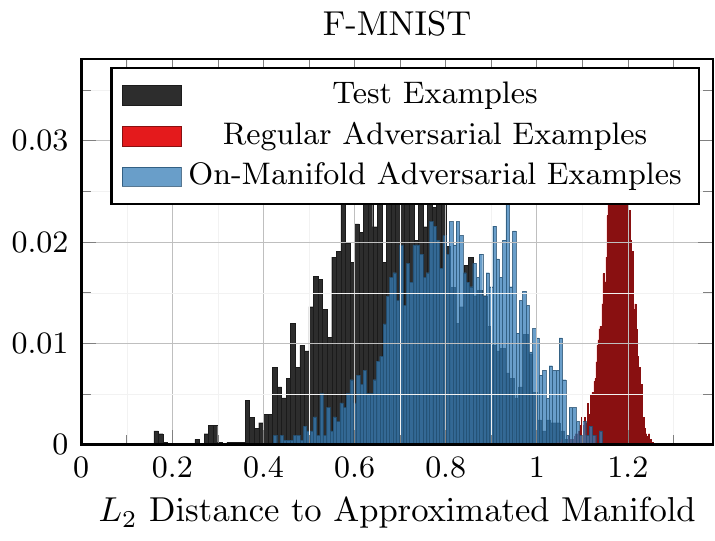}
    \end{subfigure}
    \vskip -6px
    \caption{
        On \Fonts (left), \MNIST (middle) and \Fashion (right) we plot the distance of adversarial examples to the approximated manifold. We show normalized histograms of the $L_2$ distance of adversarial examples to their projection, as described in the text. Regular adversarial examples exhibit a significant distance to the manifold; clearly distinguishable from on-manifold adversarial examples and test images. We also note that, depending on the \VAEGAN approximation, on-manifold adversarial examples are hardly distinguishable from test images.
    }
    \label{fig:appendix-off-manifold}
\end{figure*}

\section{Experimental Setup}
\label{sec:appendix-setup}

We provide technical details on the introduced synthetic \Fonts dataset, \secref{sec:appendix-fonts}, and our \VAEGAN implementation, \secref{sec:appendix-vaegan}.

\subsection{\Fonts Dataset}
\label{sec:appendix-fonts}

Our \Fonts dataset consists of randomly rotated characters ``A'' to ``J'' from different fonts, as outlined in Section 3.1 of the main paper. Specifically, we consider $1000$ Google Fonts as downloaded from the corresponding GitHub repository\footnote{\url{https://github.com/google/fonts}}. We manually exclude fonts based on symbols, or fonts that could not be rendered correctly in order to obtain a cleaned dataset consisting of clearly readable letters ``A'' to ``J''; still, the $1000$ fonts exhibit significant variance. The obtained, rendered letters are transformed using translation, shear, scaling and rotation: for each letter and font, we create $112$ transformations, uniformly sampled in $[-0.2,0.2]$, $[-0.5, 0.5]$, $[0.75,1.15]$, and $[-\nicefrac{\pi}{2},\nicefrac{\pi}{2}]$, respectively. As a result, with $1000$ fonts and $10$ classes, we obtain $1.12\text{Mio}$ images of size $28{\times}28$, splitted into $960\text{k}$ training images and $160\text{k}$ test images (of which we use $40\text{k}$ in the main paper); thus, the dataset has four times the size of \MNIST \cite{CohenARXIV2017}. For simplicity, the transformations are applied using a spatial transformer network \cite{JaderbergNIPS2015} by assembling translation $[t_1, t_2]$, shear $[\lambda_1, \lambda_2]$, scale $s$ and rotation $r$ into an affine transformation matrix,
\vskip -14px
\begin{align}
    \hskip -4px
    \left[
        \begin{matrix}
            \cos(r) s - \sin(r) s \lambda_1 & -\sin(r) s + \cos(r) s \lambda_1 & t_1\\
            \cos(r) s \lambda_2 + \sin(r) s & -\sin(r) s \lambda_2 + \cos(r) s & t_2
        \end{matrix}
    \right],
\end{align}
\vskip -4px
\noindent making the generation process fully differentiable. Overall, \Fonts offers full control over the manifold, \ie, the transformation parameters, font and class, with differentiable generative model, \ie, decoder.

\subsection{VAE-GAN Variant}
\label{sec:appendix-vaegan}

As briefly outlined in Section 3.1 of the main paper, we use class-specific \VAEGANs \cite{LarsenICML2016,RoscaARXIV2017} to approximate the class-manifolds on all datasets, \ie, \Fonts, \MNIST \cite{CohenARXIV2017}, \Fashion \cite{XiaoARXIV2017} and \Celeb \cite{LiuICCV2015}. In contrast to \cite{LarsenICML2016}, however, we use a reconstruction loss on the image, not on the discriminator's features; in contrast to \cite{RoscaARXIV2017}, we use the standard Kullback-Leibler divergence to regularize the latent space. The model consists of an encoder $\enc$, approximating the posterior $q(z|x) \approx p(z|x)$ of latent code $z$ given image $x$, a (deterministic) decoder $\dec$, and a discriminator $\dis$. During training, the sum of the following losses is minimized:
\vskip -14px
\begin{align}
    &\cL_{\enc} = \mE_{q(z|x)}\left[\lambda\|x - \dec(z)\|_1\right] + \text{KL}(q(z|x)|p(z))\label{eq:appendix-vaegan-1}\\
    &\cL_{\dec} = \mE_{q(z|x)}\left[\lambda\|x - \dec(z)\|_1 - \log(\dis(\dec(z))) \right]\label{eq:appendix-vaegan-2}\\
    &\begin{aligned}
        \cL_{\dis} = &- \mE_{p(x)}\left[\log(\dis(x))\right]\\
        &- \mE_{q(z|x)}\left[\log(1 - \dis(\dec(z)))\right]\label{eq:appendix-vaegan-3}
    \end{aligned}
\end{align}
\vskip -4px
\noindent using a standard Gaussian prior $p(z)$. Here, $q(z|x)$ is modeled by predicting the mean $\mu(x)$ and variance $\sigma^2(x)$ such that $q(z|x) = \mathcal{N}(z; \mu(x), \text{diag}(\sigma^2(x)))$ and the weighting parameter $\lambda$ controls the importance of the $L_1$ reconstruction loss relative to the Kullback-Leibler divergence $\text{KL}$ and the adversarial loss for decoder and discriminator. As in~\cite{KingmaICLR2014}, we use the reparameterization trick with one sample to approximate the expectations in \eqnref{eq:appendix-vaegan-1}, \eqref{eq:appendix-vaegan-2} and \eqref{eq:appendix-vaegan-3}, and the Kullback-Leibler divergence $\text{KL}(q(z|x)|p(z))$ is computed analytically.

The encoder, decoder and discriminator consist of three (four for \Celeb) (de-) convolutional layers ($4{\times}4$ kernels; stride $2$; $64$, $128$, $256$ channels), followed by ReLU activations and batch normalization \cite{IoffeICML2015}; the encoder uses two fully connected layers to predict mean and variance; the discriminator uses two fully connected layers to predict logits. We tuned $\lambda$ to dataset- and class-specific values: on \Fonts, $\lambda = 3$ worked well for all classes, on \MNIST, $\lambda = 2.5$ except for classes ``0'' ($\lambda = 2.75$), ``1'' ($\lambda = 5.6$) and ``8'' ($\lambda = 2.25$), on \Fashion, $\lambda = 2.75$ worked well for all classes, on \Celeb $\lambda = 3$ worked well for both classes. Finally, we trained our \VAEGANs using ADAM~\cite{KingmaICLR2015} with learning rate $0.005$ (decayed by $0.9$ every epoch), weight decay $0.0001$ and batch size $100$ for $10$, $30$, $60$ and $30$ epochs on \Fonts, \MNIST, \Fashion and \Celeb, respectively. We also consider class-agnostic \VAEGANs trained using the same strategy with $\lambda = 3$ for \Fonts, $\lambda = 3$ on \MNIST, $\lambda =2.75$ on \Fashion and $\lambda=3$ on \Celeb, see \secref{sec:appendix-data-manifold} for results.

In \figref{fig:appendix-vaegan}, we include random samples of the class-specific \VAEGANs. Especially on \MNIST and \Fonts, our \VAEGANs generate realistic looking samples with sharp edges. However, we also show several problematic random samples, illustrating the discrepancy between the true data distribution and the approximation -- as particularly highlighted on \Fonts.

\section{Adversarial Example Distance to Manifold}
\label{sec:appendix-off-manifold}

Complementing Section 3.2 of the main paper, we provide additional details and results regarding the distance of regular adversarial examples to the true or approximated manifold, \red{including a theoretical argument of adversarial examples leaving the manifold}.

\begin{figure*}[t]
    \centering
    \vskip -0.4cm
    \includegraphics[width=1\textwidth]{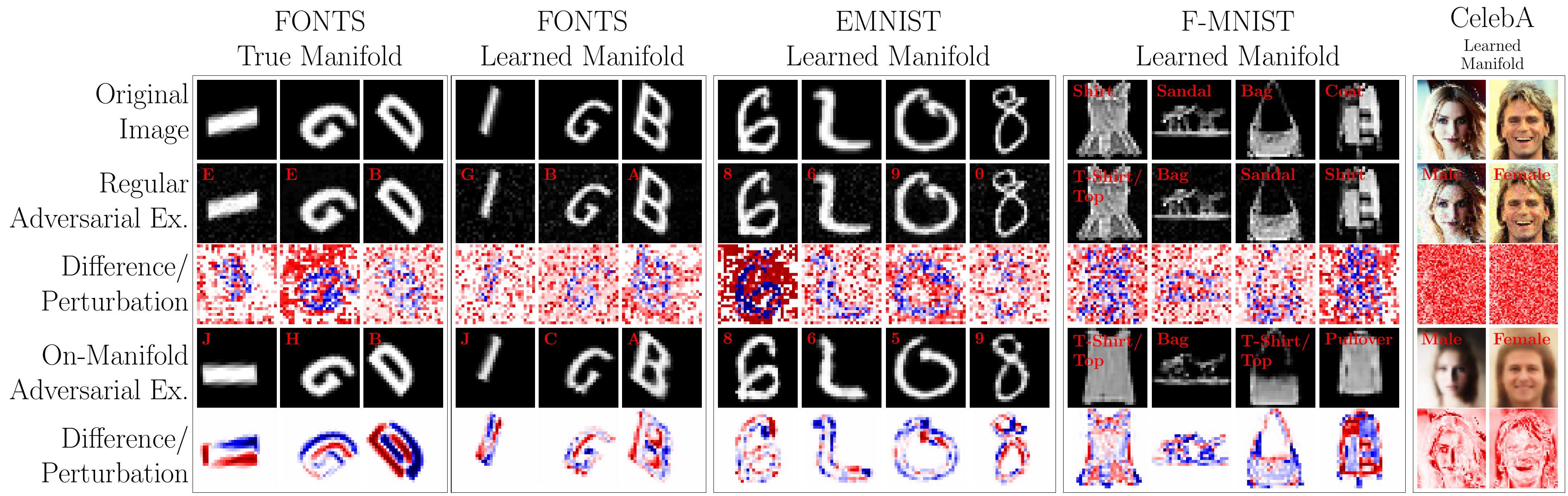}
    \vskip -6px 
    \caption{Regular and on-manifold adversarial examples on \Fonts, \MNIST, \Fashion and \Celeb. On \Fonts, the manifold is known; on the other datasets, class manifolds have been approximated using \VAEGANs. Notice that the crafted on-manifold adversarial examples correspond to meaningful manipulations of the image -- as long as the learned class-manifolds are good approximations. This can best be seen considering the (normalized) difference images (or the magnitude thereof for \Celeb).}
    \label{fig:appendix-examples}
\end{figure*}

On \Fonts, with access to the true manifold in form of a perfect decoder $\dec$, we iteratively obtain the latent code $\tilde{z}$ yielding the manifold's closest image to the given adversarial example $\tilde{x}$ as
\vskip -14px
\begin{align}
    \tilde{z} = \argmin_z \|\tilde{x} - \dec(z)\|_2^2.
\end{align}
\vskip -4px
\noindent We use $100$ iterations of ADAM \cite{KingmaICLR2015}, with a learning rate of $0.09$, decayed every $10$ iterations by a factor $0.95$. We found that additional iterations did not improve the results. The obtained projection $\pi(\tilde{x}) = \dec(\tilde{z})$ is usually very close to the original test image $x$ for which the adversarial example was crafted. The distance is then computed as $\|\tilde{x} - \pi(\tilde{x})\|_2$; we refer to the main paper for results and discussion.

If the true manifold is not available, we locally approximate the manifold using $50$ nearest neighbors $x_1,\ldots,x_{50}$ of the adversarial example $\tilde{x}$. In the main paper, we center these nearest neighbors at the test image $x$, \ie, consider the sub-space spanned by $x_i - x$. Here, we show that the results can be confirmed when centering the nearest neighbors at their mean $\bar{x} = \nicefrac{1}{50} \sum_{i = 1}^{50} x_i$ and considering the subspace spanned by $x_i - \bar{x}$ instead. In this scenario, the test image $x$ is not necessarily part of the approximated manifold anymore. The projection onto this sub-space can be obtained by solving the least squares problem; specifically, we consider the vector $\delta = \tilde{x} - x$, \ie, we assume that the ``adversarial direction'' originates at the mean $\bar{x}$. Then, we solve
\vskip -14px
\begin{align}
    \beta^\ast = \argmin_\beta \|X\beta - \delta\|_2^2\label{eq:appendix-least-squares}
\end{align}
\vskip -4px
\noindent where the columns $X_i$ are the vectors $x_i - \bar{x}$. The projection $\pi(\tilde{x})$ is obtained as $\pi(\tilde{x}) = X\beta^\ast$; the same approach can be applied to projecting the test image $x$. Note that it is crucial to consider the adversarial direction $\delta$ itself, instead of the adversarial example~$\tilde{x}$ because $\|\delta\|_2$ is small by construction, \ie, the projections of $\tilde{x}$ and $x$ are very close. In \figref{fig:appendix-off-manifold}, we show results using this approximation on \Fonts, \MNIST and \Fashion. Regular adversarial examples can clearly be distinguished from test images and on-manifold adversarial examples. Note, however, that we assume access to both the test image $x$ and the corresponding adversarial example $\tilde{x}$ such that this finding cannot be exploited for detection. We also notice that the discrepancy between the distance distributions of test images and on-manifold adversarial examples reflects the approximation quality of the used \VAEGANs.

\subsection{Intuition and Theoretical Argument}

\red{Having empirically shown that regular adversarial examples tend to leave the manifold, often in a nearly orthogonal direction, we also discuss a theoretical argument supporting this observation. The main assumption is that the training loss is constant on the manifold (normally close to zero) due to training and proper generalization, \ie, low training and test loss. Thus, the loss gradient is approximately orthogonal to the manifold as this is the direction to increase the loss most efficiently.}

\begin{figure*}[t]
    \centering
    \vskip -0.4cm
    \begin{subfigure}{0.255\textwidth}
        \centering
        \includegraphics[width=\textwidth]{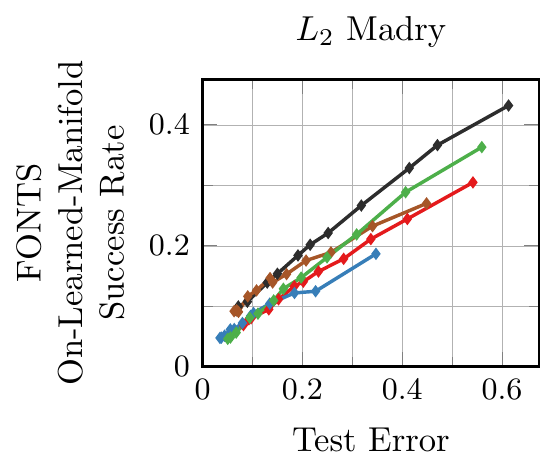}
    \end{subfigure}
    \begin{subfigure}{0.235\textwidth}
        \centering
        \includegraphics[width=\textwidth]{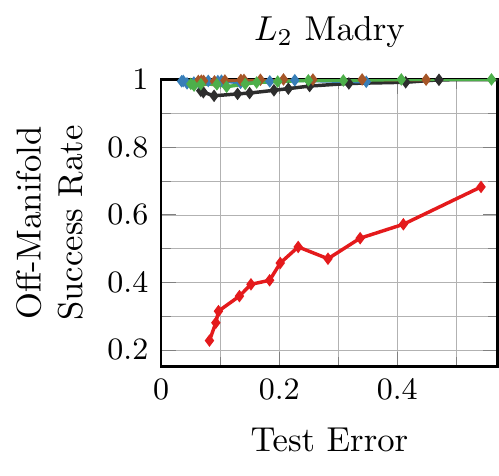}
    \end{subfigure}
    \begin{subfigure}{0.235\textwidth}
        \centering
        \includegraphics[width=\textwidth]{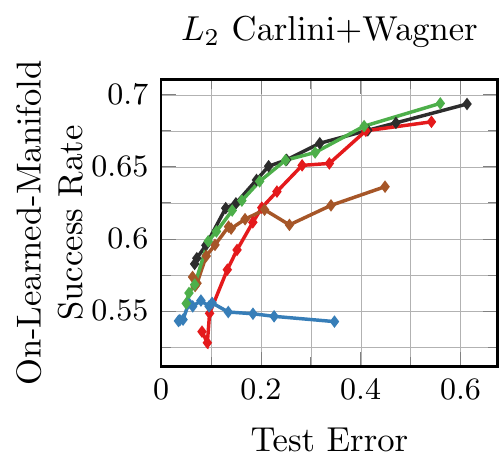}
    \end{subfigure}
    \begin{subfigure}{0.235\textwidth}
        \centering
        \includegraphics[width=\textwidth]{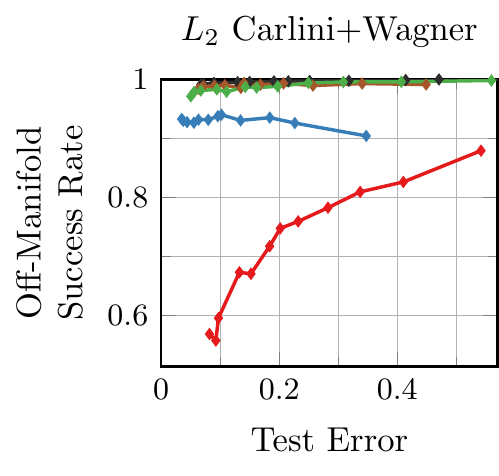}
    \end{subfigure}
    \\
    \begin{subfigure}{0.255\textwidth}
        \centering
        \includegraphics[width=\textwidth]{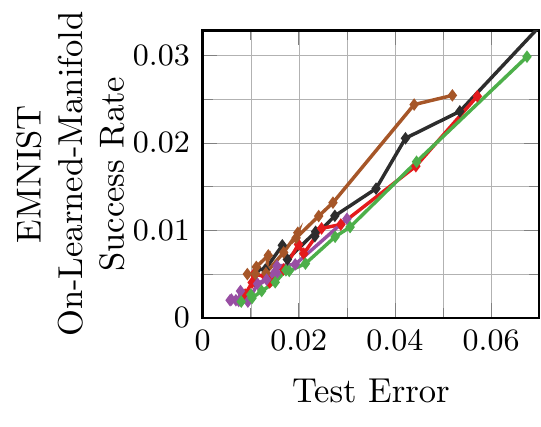}
    \end{subfigure}
    \begin{subfigure}{0.235\textwidth}
        \centering
        \includegraphics[width=\textwidth]{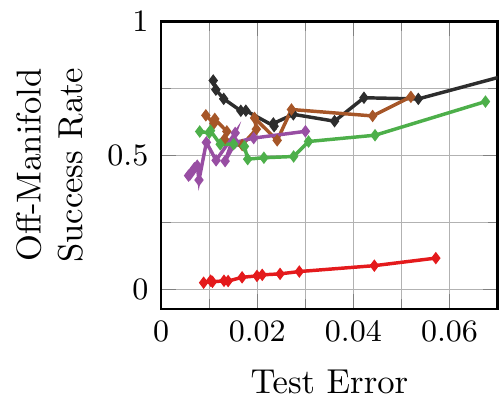}
    \end{subfigure}
    \begin{subfigure}{0.235\textwidth}
        \centering
        \includegraphics[width=\textwidth]{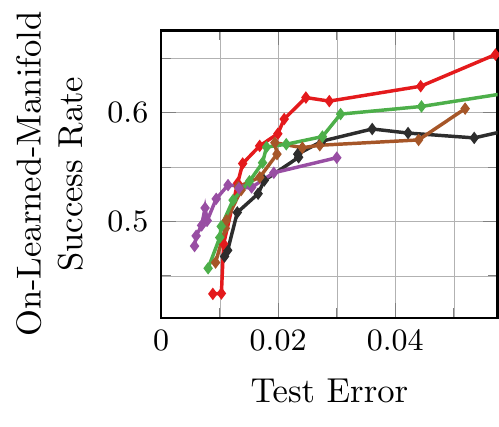}
    \end{subfigure}
    \begin{subfigure}{0.235\textwidth}
        \centering
        \includegraphics[width=\textwidth]{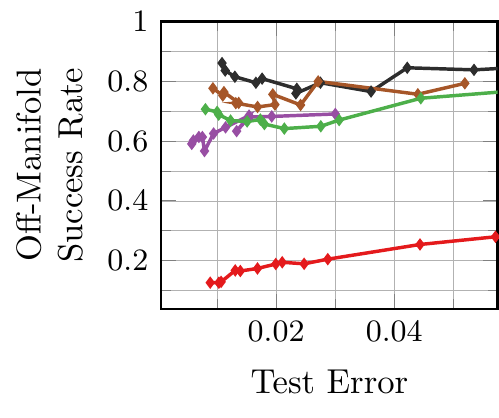}
    \end{subfigure}
    \\
    \begin{subfigure}{0.255\textwidth}
        \centering
        \includegraphics[width=\textwidth]{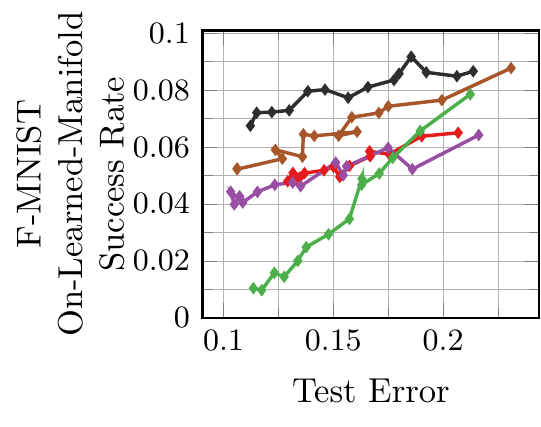}
    \end{subfigure}
    \begin{subfigure}{0.235\textwidth}
        \centering
        \includegraphics[width=\textwidth]{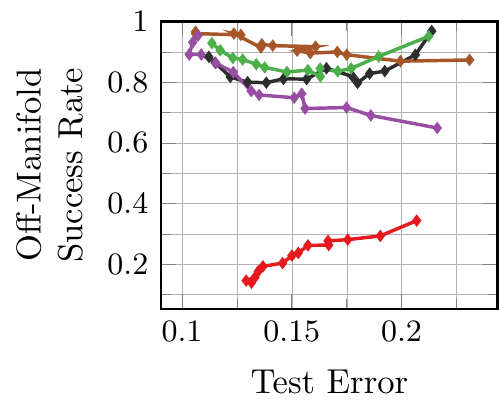}
    \end{subfigure}
    \begin{subfigure}{0.235\textwidth}
        \centering
        \includegraphics[width=\textwidth]{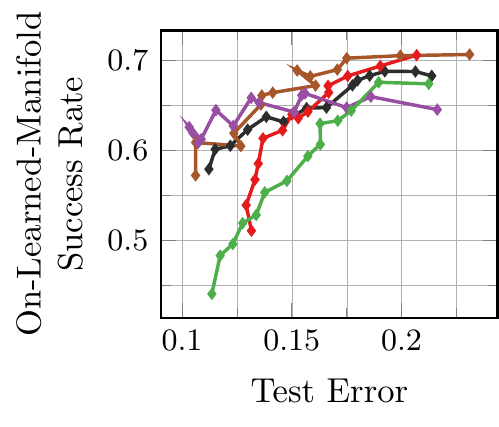}
    \end{subfigure}
    \begin{subfigure}{0.235\textwidth}
        \centering
        \includegraphics[width=\textwidth]{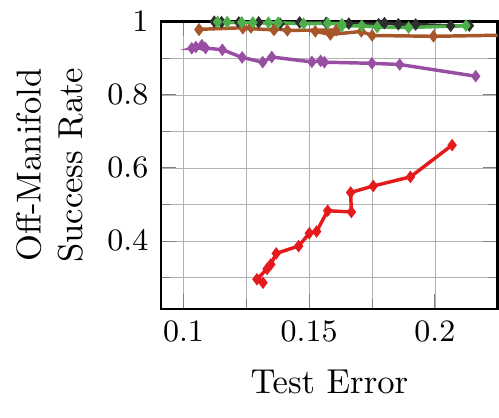}
    \end{subfigure}
    \\
    \fcolorbox{black!50}{white}{
        \begin{subfigure}{1\textwidth}
            \centering
            \includegraphics[width=0.825\textwidth]{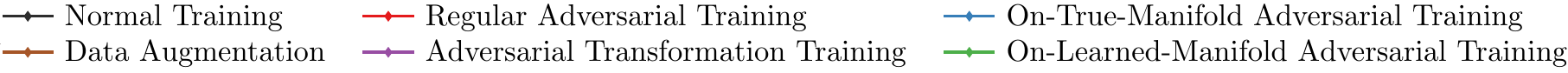}
        \end{subfigure}
    }
    \caption{
        $L_2$ attacks of Madry \etal \cite{MadryICLR2018} and Carlini and Wagner \cite{CarliniSP2017} on \Fonts, \MNIST and \Fashion. In all cases, we plot regular or on-manifold success rate against test error. Independent of the attack, we can confirm that on-manifold robustness is strongly related to generalization, while regular robustness is independent of generalization.
    }
    \label{fig:appendix-l2}
\end{figure*}

\red{More formally, let $f(x)$ denote the classifier which -- for simplicity -- takes inputs $x \in \mathbb{R}^d$ and predicts outputs $y \in \mathbb{R}^K$ for $K$ classes. We assume both the classifier as well as the used loss, \eg, cross-entropy loss, to be differentiable. We further expect the data to lie on a manifold $\mathcal{M}$ and the loss to be constant on $\mathcal{M} \cap B(x, \epsilon)$ with 
\begin{align}
    B(x, \epsilon) = \{x' \in \mathbb{R}^d: \|x' - x\| \leq \epsilon\}.
\end{align}
Let
\begin{align}
    g(x) = \mathbb{E}\left[ \cL(f(x), y)|x\right]
\end{align}
be the conditional expectation of the loss $\cL$; then, by the mean value theorem, there exists $\theta(x') \in [0,1]$ for each $x' \in \mathcal{M} \cap B(x, \epsilon)$ such that
\begin{align}
    0 &= g(x') - g(x)\\
    &= \inner{\nabla g(\theta(x') x + (1 - \theta(x'))x'), x' - x}
\end{align}
As this holds for all $\epsilon > 0$ and as $\epsilon \rightarrow 0$, every vector $x' - x$ becomes a tangent of $\mathcal{M}$ at $x$ and
\begin{align}
    \lim_{\epsilon \rightarrow 0} \nabla g(\theta(x') x + (1 - \theta(x'))x') = \nabla g(x),
\end{align}
it holds that $\nabla g(x)$ is orthogonal to the tangent space of $\mathcal{M}$ at $x$. As $\nabla g(x)$ is the gradient of the expected loss, it implies that adversarial examples, as computed, \eg, using first-order gradient-based approaches such as \eqnref{eq:appendix-off-manifold-madry}, leave the manifold $\mathcal{M}$ in order to fool the classifier $f(x)$.}

\section{On-Manifold Adversarial Examples}
\label{sec:appendix-on-manifold}

\begin{figure*}[t]
    \centering
    \vskip -0.4cm
    \includegraphics[width=1\textwidth]{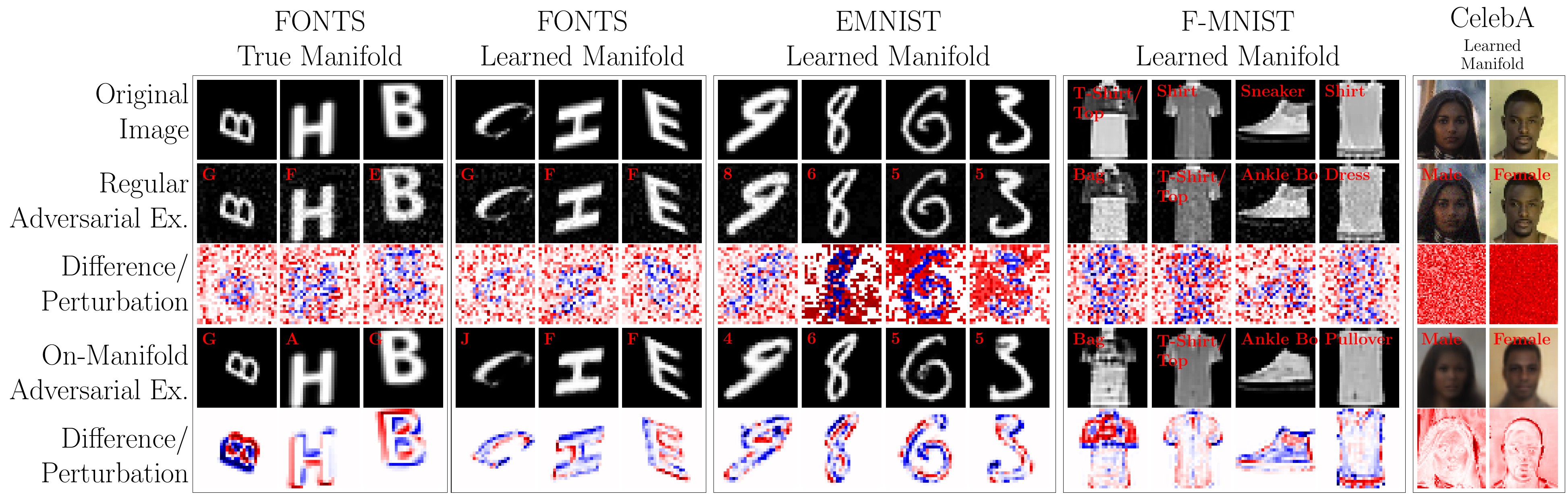}
    \vskip -6px 
    \caption{Regular and on-manifold adversarial examples on \Fonts, \MNIST, \Fashion and \Celeb. On \Fonts, the manifold is known; otherwise, class manifolds have been approximated using \VAEGANs. In addition to the original test images, we also show the adversarial examples and their (normalized) difference (or the magnitude thereof for \Celeb).}
    \label{fig:appendix-examples-2}
\end{figure*}

In \figref{fig:appendix-examples}, we show additional examples of regular and on-manifold adversarial examples, complementing the examples in Fig.\ 2 of the main paper. On \Fonts, both using the true and the approximated manifold, on-manifold adversarial examples reflect the underlying invariances of the data, \ie, the transformations employed in the generation process. This is in contrast to the corresponding regular adversarial examples and their (seemingly) random noise patterns. We note that regular and on-manifold adversarial examples can best be distinguished based on their difference to the original test image -- although both are perceptually close to the original image. Similar observations hold on \MNIST and \Fashion. However, especially on \Fashion and \Celeb, the discrepancy between true images and on-manifold adversarial examples becomes visible. This is the ``cost'' of approximating the underlying manifold using \VAEGANs. More examples can be found in \figref{fig:appendix-examples-2} at the end of this document.

\section{$L_2$ and Transfer Attacks}
\label{sec:appendix-attacks}

\begin{figure}[t]
    \centering
    \vskip -0.4cm
    \begin{subfigure}{0.245\textwidth}
        \centering
        \includegraphics[width=\textwidth]{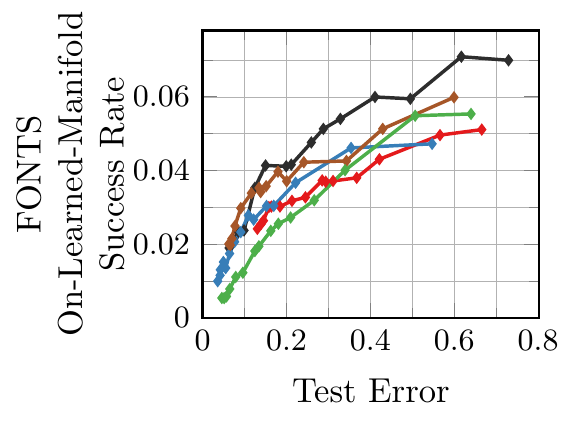}
    \end{subfigure}
    \begin{subfigure}{0.225\textwidth}
        \centering
        \includegraphics[width=\textwidth]{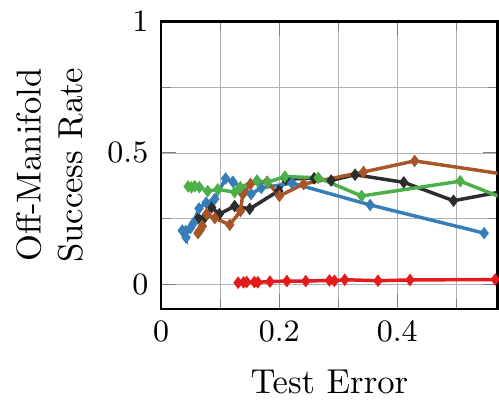}
    \end{subfigure}
    \\
    \begin{subfigure}{0.245\textwidth}
        \centering
        \includegraphics[width=\textwidth]{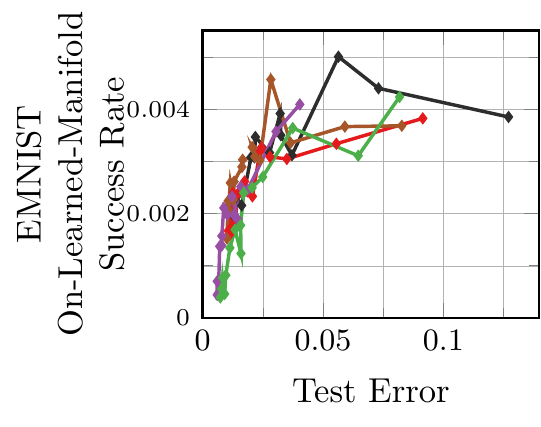}
    \end{subfigure}
    \begin{subfigure}{0.225\textwidth}
        \centering
        \includegraphics[width=\textwidth]{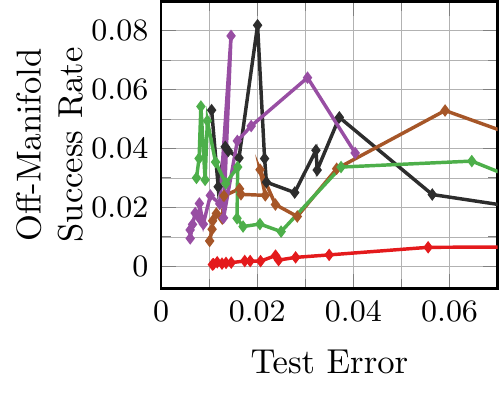}
    \end{subfigure}
    \\
    \begin{subfigure}{0.245\textwidth}
        \centering
        \includegraphics[width=\textwidth]{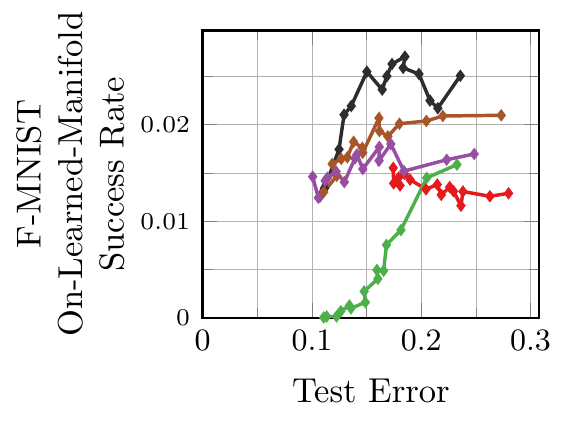}
    \end{subfigure}
    \begin{subfigure}{0.225\textwidth}
        \centering
        \includegraphics[width=\textwidth]{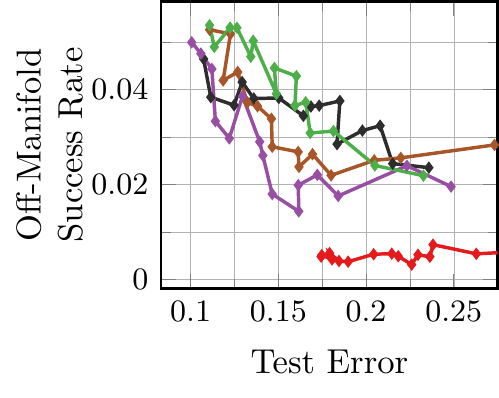}
    \end{subfigure}
    \\
    \fcolorbox{black!50}{white}{
        \begin{subfigure}{0.45\textwidth}
            \centering
            \includegraphics[width=1\textwidth]{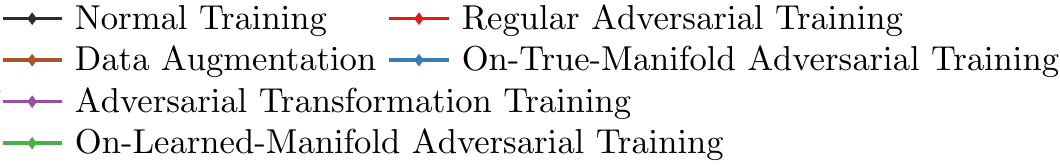}
        \end{subfigure}
    }
    \caption{Transfer attacks on \Fonts, \MNIST and \Fashion. We show on-manifold (left) and regular success rate (right) plotted against test error. In spite of significantly lower success rates, transfer attacks also allow to confirm the strong relationship between on-manifold success rate and test error, while -- at least on \Fonts and \MNIST -- regular success rate is independent of test error.}
    \label{fig:appendix-transfer}
\end{figure}

In the main paper, see Section 3.1, we primarily focus on the $L_{\infty}$ white-box attack by Madry \etal \cite{MadryICLR2018}. Here, we further consider the $L_2$ variant, which, given image $x$ with label $y$ and classifier $f$, maximizes the training loss, \ie,
\vskip -14px
\begin{align}
    \max_\delta \cL(f(x + \delta), y)\text{ s.t. }\|\delta\|_2 \leq \epsilon, \tilde{x}_i \in [0,1],
    \label{eq:appendix-off-manifold-madry}
\end{align}
\vskip -4px
\noindent to obtain an adversarial example $\tilde{x} = x + \delta$. \red{We use $\epsilon = 1.5$ for regular adversarial examples and $\epsilon = 0.3$ for on-manifold adversarial examples. For optimization, we utilize} projected ADAM \cite{KingmaICLR2015}: after each iteration, $\tilde{x}$ is projected onto the $L_2$-ball of radius $\epsilon$ using 
\vskip -14px
\begin{align}
    \tilde{x}' = \tilde{x} \cdot \max\left(1, \frac{\epsilon}{\|\tilde{x}\|_2}\right)
\end{align}
\vskip -4px
\noindent and clipped to $[0, 1]$. We use a learning rate of $0.005$ and we note that ADAM includes momentum, as suggested in~\cite{DongCVPR2018}. Optimization stops as soon as the label changes, or runs for a maximum of $40$ iterations. The perturbation $\delta$ is initialized randomly as follows:
\vskip -14px
\begin{align}
    \delta = u \epsilon \frac{\delta'}{\|\delta'\|_2},\quad \delta' \sim \mathcal{N}(0, I), u \sim U(0,1).
\end{align}
\vskip -4px
\noindent Here, $U(0,1)$ refers to the uniform distribution over $[0,1]$. This results in $\delta$ being in the $\epsilon$-ball and uniformly distributed over distance and direction. Note that this is in contrast to sampling uniformly \wrt the volume of the $\epsilon$-ball. The same procedure applies to the $L_{\infty}$ attack where the projection onto the $\epsilon$-ball is achieved by clipping. The attack can also be used to obtain on-manifold adversarial examples, as described in Section 3.3 of the main paper. Then, optimization in \eqnref{eq:appendix-off-manifold-madry} is done over the perturbation $\zeta$ in latent space, with constraint $\|\zeta\|_2 \leq \eta$. The adversarial example is obtained as $\tilde{x} = \dec(z + \zeta)$ with $z$ being the latent code of image $x$ and $\dec$ being the true or approximated generative model, \ie, decoder.

\begin{figure}[t]
    \centering
    \vskip -0.4cm
    \hskip -0.1cm
    \begin{subfigure}[t]{0.5\textwidth}
        \begin{subfigure}[t]{0.49\textwidth}
            \includegraphics[width=0.9\textwidth]{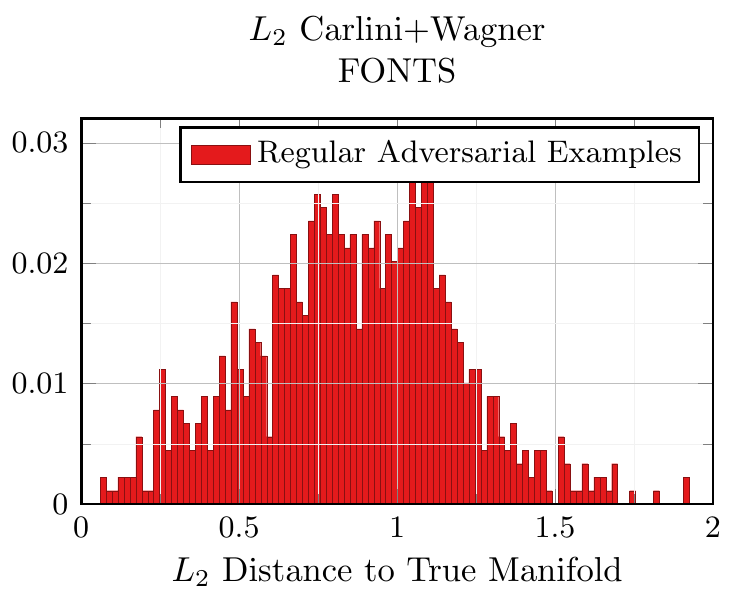}
        \end{subfigure}
        \begin{subfigure}[t]{0.49\textwidth}
            \includegraphics[width=0.9\textwidth]{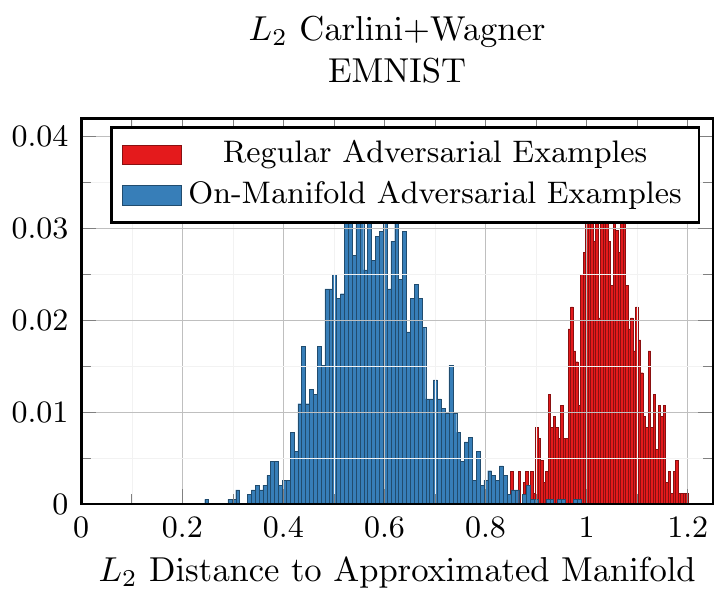}
        \end{subfigure}
    \end{subfigure}
    \vskip -6px
    \caption{\red{Distance of Carlini+Wagner adversarial examples to the true, on \Fonts (left), or approximated, on \MNIST (right), manifold. As before, we show normalized histograms of the $L_2$ distance of adversarial examples to their projections onto the manifold. Even for different attacks and the $L_2$ norm, regular adversarial examples seem to leave the manifold.}}
    \label{fig:appendix-off-manifold-cw}
\end{figure}

We also consider the $L_2$ white box attack by Carlini and Wagner \cite{CarliniSP2017}. Instead of directly maximizing the training loss, Carlini and Wagner propose to use a surrogate objective on the classifier's logits $l_y$:
\vskip -14px
\begin{align}
    F(\tilde{x}, y) &= \max(-\kappa, l_y(\tilde{x}) - \max_{y' \neq y} l_{y'}(\tilde{x})).
\end{align}
\vskip -4px
\noindent Compared to the training loss, which might be close to zero for a well-trained network, $F$ is argued to provide more useful gradients \cite{CarliniSP2017}. Then,
\vskip -14px
\begin{align}
    \min_\delta F(x + \delta, y) + \lambda \|\delta\|_2\text{ s.t. }\tilde{x}_i \in [0,1]
\end{align}
\vskip -4px
\noindent is minimized by reparameterizing $\delta$ in terms of $\delta= \nicefrac{1}{2}(\tanh(\omega) + 1)- x$ in order to ensure the image-constraint, \ie, $\tilde{x}_i \in [0,1]$. In practice, we empirically chose $\kappa = 1.5$, use $120$ iterations of ADAM \cite{KingmaICLR2015} with learning rate $0.005$ and $\lambda = 1$. Again, this attack can be used to obtain on-manifold adversarial examples, as well.

As black-box attack we transfer $L_\infty$ Madry adversarial examples from a held out model, as previously done in \cite{LiuICLR2017,XieARXIV2018,PapernotASIACCS2017}. The held out transfer model is trained normally, \ie, without any data augmentation or adversarial training, on $10\text{k}$ training images for $20$ epochs (as outlined in Section 3.1 of the main paper). The success rate of these transfer attacks is computed with respect to images that are correctly classified by both the transfer model and the target model.

Extending the discussion of Sections 3.4 and 3.5 of the main paper, \figref{fig:appendix-l2} shows results on \Fonts, \MNIST and \Fashion considering both $L_2$ attacks, \ie, Madry \etal \cite{MadryICLR2018} and Carlini and Wagner \cite{CarliniSP2017}. In contrast to the $L_{\infty}$ Madry attack, we observe generally lower success rates. Nevertheless, we can observe a clear relationship between on-manifold success rate and test error. The exact form of this relationship, however, depends on the attack; for the $L_2$ Madry attack, the relationships seems to be mostly linear (especially on \Fonts and \MNIST), while it seems non-linear for the $L_2$ Carlini and Wagner attack. Furthermore, the independence of regular robustness and generalization can be confirmed, \ie, regular success rate is roughly constant when test error varies -- again, with the exception of regular adversarial training. \red{Finally, for completeness, in \figref{fig:appendix-off-manifold-cw}, we illustrate that the Carlini+Wagner $L_2$ attack also results in regular adversarial examples leaving the manifold.}

\begin{figure}[t]
    \centering
    \vskip -0.4cm
    \begin{subfigure}{0.245\textwidth}
        \centering
        \includegraphics[width=\textwidth]{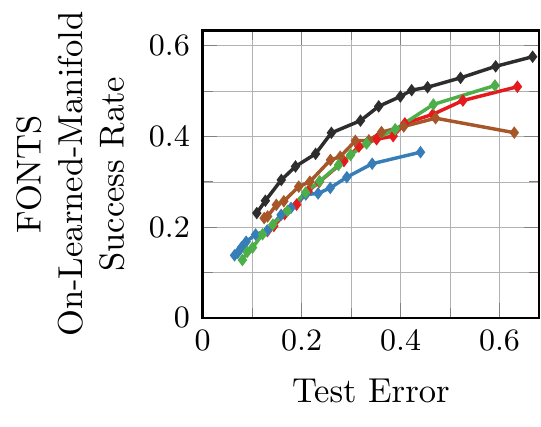}
    \end{subfigure}
    \begin{subfigure}{0.225\textwidth}
        \centering
        \includegraphics[width=\textwidth]{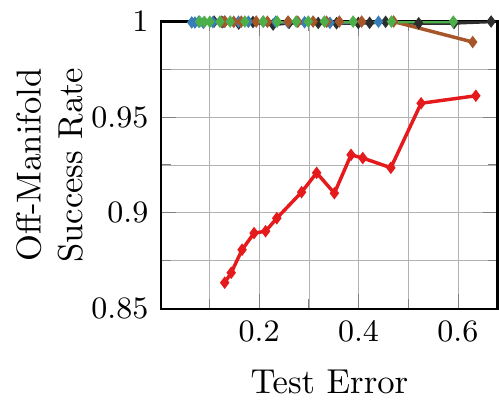}
    \end{subfigure}
    \\
    \begin{subfigure}{0.245\textwidth}
        \centering
        \includegraphics[width=\textwidth]{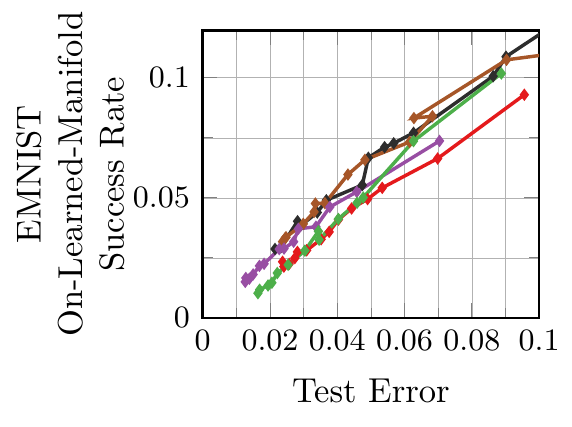}
    \end{subfigure}
    \begin{subfigure}{0.225\textwidth}
        \centering
        \includegraphics[width=0.925\textwidth]{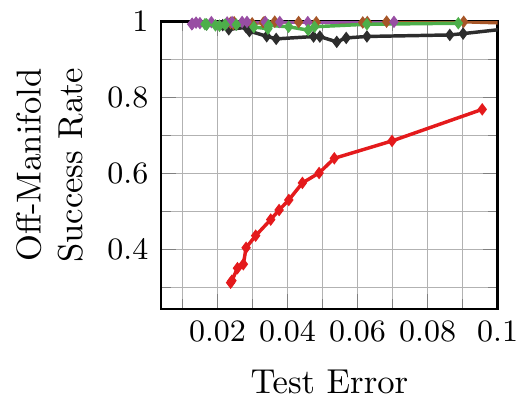}
    \end{subfigure}
    \\
    \begin{subfigure}{0.245\textwidth}
        \centering
        \includegraphics[width=\textwidth]{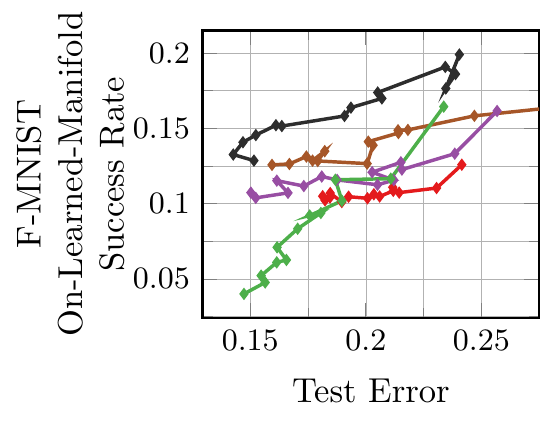}
    \end{subfigure}
    \begin{subfigure}{0.225\textwidth}
        \centering
        \includegraphics[width=0.925\textwidth]{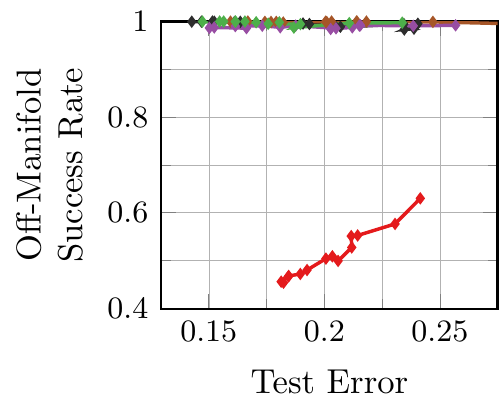}
    \end{subfigure}
    \\
    \fcolorbox{black!50}{white}{
        \begin{subfigure}{0.45\textwidth}
            \centering
            \includegraphics[width=1.01\textwidth]{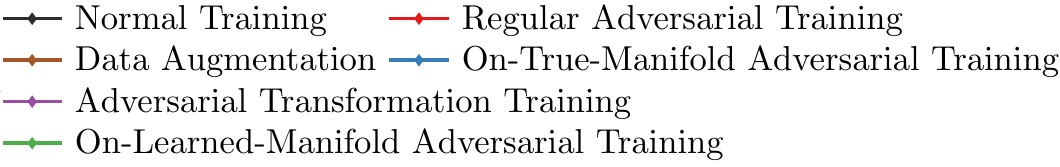}
        \end{subfigure}
    }
    \caption{Experiments with multilayer-perceptrons on \Fonts, \MNIST and \Fashion. We plot on-manifold (left) or regular success rate (right) against test error. On-manifold robustness is strongly related to generalization, while regular robustness seems mostly independent of generalization.}
    \label{fig:appendix-mlp}
\end{figure}
\begin{figure}
    \centering
    \vskip -0.4cm
    \begin{subfigure}{0.245\textwidth}
        \centering
        \includegraphics[width=\textwidth]{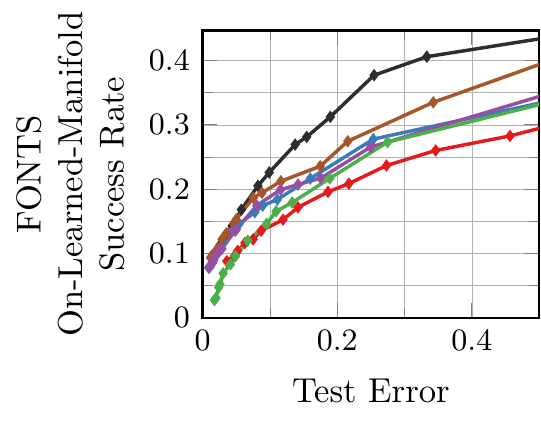}
    \end{subfigure}
    \begin{subfigure}{0.225\textwidth}
        \centering
        \includegraphics[width=\textwidth]{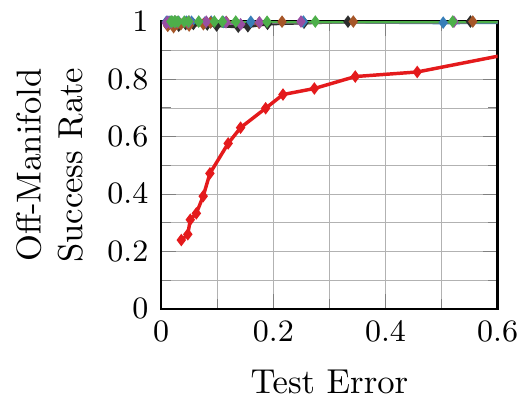}
    \end{subfigure}
    \\
    \begin{subfigure}{0.245\textwidth}
        \centering
        \includegraphics[width=\textwidth]{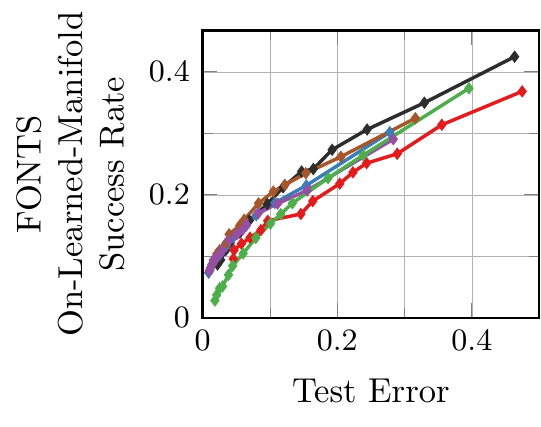}
    \end{subfigure}
    \begin{subfigure}{0.225\textwidth}
        \centering
        \includegraphics[width=0.925\textwidth]{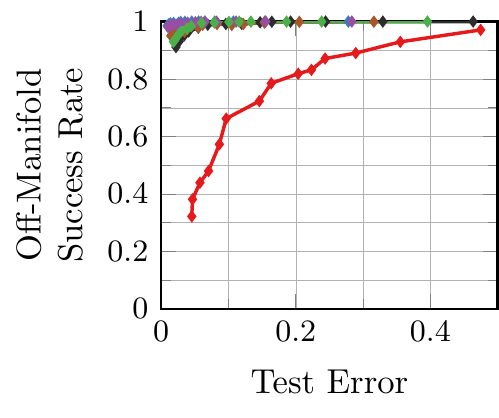}
    \end{subfigure}
    \\
    \begin{subfigure}{0.245\textwidth}
        \centering
        \includegraphics[width=\textwidth]{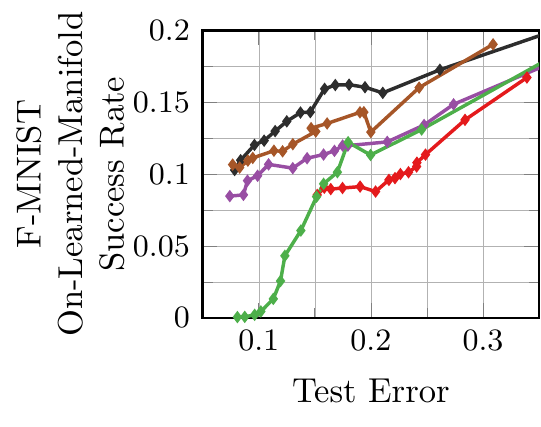}
    \end{subfigure}
    \begin{subfigure}{0.225\textwidth}
        \centering
        \includegraphics[width=\textwidth]{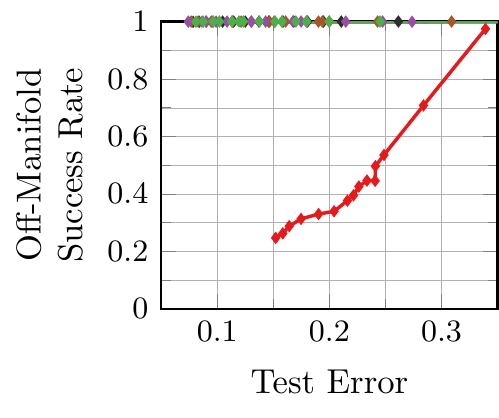}
    \end{subfigure}
    \\
    \begin{subfigure}{0.245\textwidth}
        \centering
        \includegraphics[width=\textwidth]{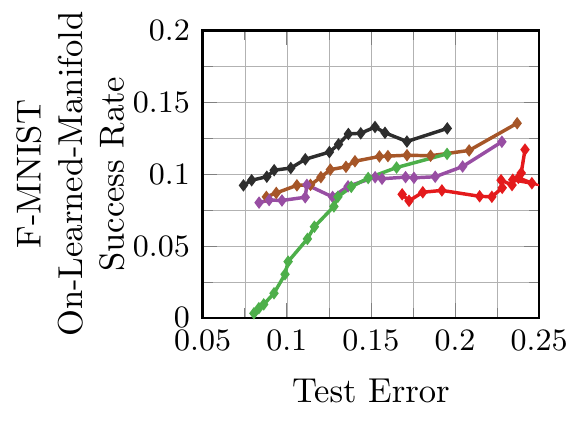}
    \end{subfigure}
    \begin{subfigure}{0.225\textwidth}
        \centering
        \includegraphics[width=0.925\textwidth]{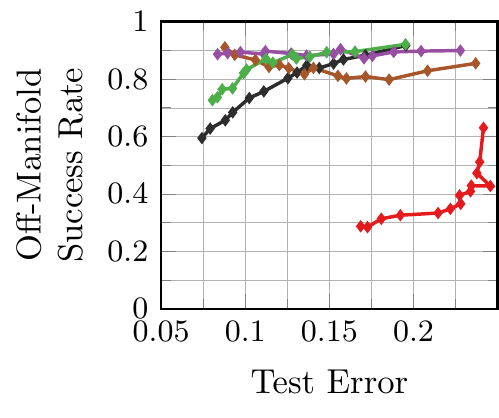}
    \end{subfigure}
    \\
    \fcolorbox{black!50}{white}{
        \begin{subfigure}{0.45\textwidth}
            \centering
            \includegraphics[width=1.01\textwidth]{appendix_mlp_legend.pdf}
        \end{subfigure}
    }
    \caption{\red{Experiments with ResNet-13 (top) and VGG (bottom) on \Fonts and \Fashion. We plot on-manifold (left) or regular success rate (right) against test error. As in \figref{fig:appendix-mlp}, our claims can be confirmed for these network architectures, as well.}}
    \label{fig:appendix-resnet-vgg}
\end{figure}

In \figref{fig:appendix-transfer}, we also consider the black-box case, \ie, without access to the target model. While both observations from above can be confirmed, especially on \Fonts and \MNIST, the results are significantly less pronounced. This is mainly due to the significantly lower success rate of transfer attacks -- both regarding regular and on-manifold adversarial examples. Especially on \MNIST and \Fashion, success rate may reduce from previously $80\%$ or higher to $10\%$ or lower. This might also explain the high variance on \MNIST and \Fashion regarding regular robustness. Overall, we demonstrate that our claims can be confirmed in both white- and black-box settings as well as using different attacks \cite{MadryICLR2018,CarliniSP2017} and norms.

\section{Influence of Network Architecture}
\label{sec:appendix-network}

Also in relation to the discussion in Sections 3.4 and 3.5 of the main paper, \figref{fig:appendix-mlp} shows results on \Fonts, \MNIST and \Fashion using multi-layer perceptrons instead of convolutional neural networks. Specifically, we consider a network with $4$ hidden layers, using $128$ hidden units each; each layer is followed by ReLU activations and batch normalization \cite{IoffeICML2015}; training strategy, however, remains unchanged. Both of our claims, \ie, that on-manifold robustness is essentially generalization but regular robustness is independent of generalization, can be confirmed. Especially regarding the latter, results are more pronounced using multi-layer perceptrons: except for regular adversarial training, success rate stays nearly constant at $100\%$ irrespective of test error. Overall, these results suggest that our claims generally hold for the class of (deep) neural networks, irrespective of architectural details.

\red{In order to further validate our claims, we also consider variants of two widely used, state-of-the-art architectures: ResNet-13 \cite{HeCVPR2016} and VGG \cite{SimonyanARXIV2014}. For VGG, however, we removed the included dropout layers. The main reason is that randomization might influence robustness, \eg, see \cite{AthalyeARXIV2018}. Additionally, we only use 2 stages of model A, see \cite{SimonyanARXIV2014}, in order to deal with the significantly lower resolution of $28 \times 28$ on \Fonts, \MNIST and \Fashion; finally, we only use $1024$ hidden units in the fully connected layers. \figref{fig:appendix-resnet-vgg} shows results on \Fonts and \Fashion (which are significantly more difficult than \MNIST) confirming our claims.}

\section{From Class Manifolds to Data Manifold}
\label{sec:appendix-data-manifold}

\begin{figure}[t]
    \centering
    \vskip -0.4cm
    \includegraphics[width=0.45\textwidth]{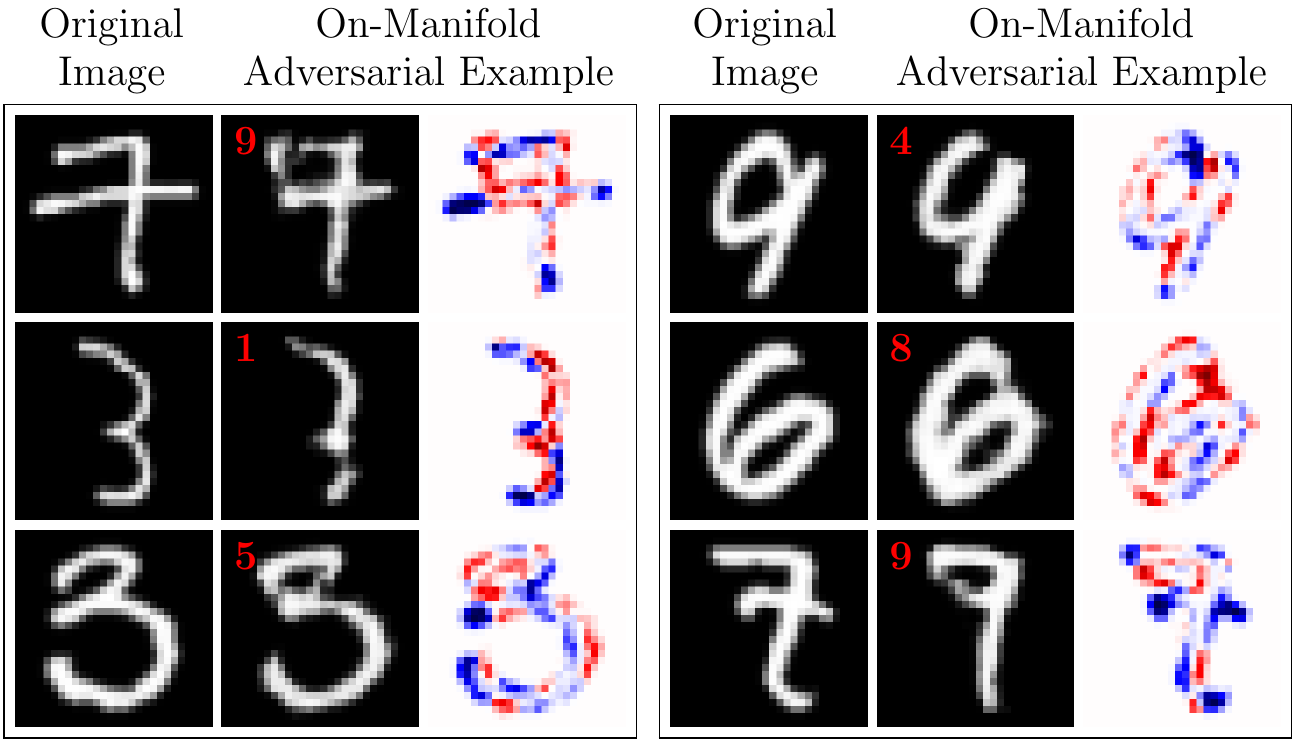}
    \caption{On-manifold adversarial examples crafted using class-agnostic \VAEGANs on \MNIST. We show examples illustrating the problematic of unclear class boundaries within the learned manifold. On-manifold adversarial examples are not guaranteed to be label invariant, \ie, they may change the actual, true label according to the approximate data distribution.}
    \label{fig:appendix-data-examples}
\end{figure}
\begin{figure}[t]
    \centering
    \vskip -0.4cm
    \begin{subfigure}{0.235\textwidth}
        \centering
        \includegraphics[width=\textwidth]{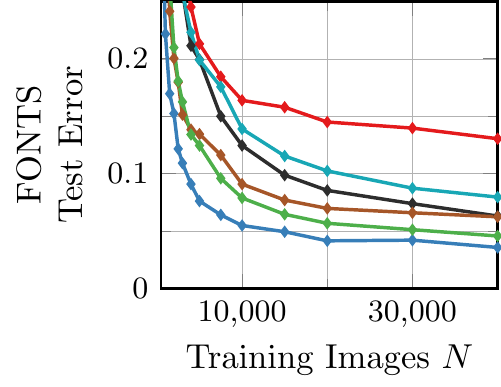}
    \end{subfigure}
    \begin{subfigure}{0.235\textwidth}
        \centering
        \includegraphics[width=\textwidth]{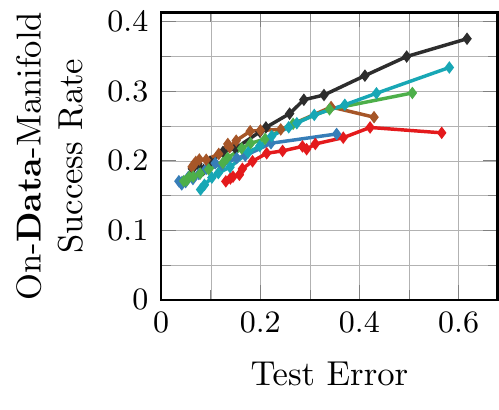}
    \end{subfigure}
    \\
    \begin{subfigure}{0.235\textwidth}
        \centering
        \includegraphics[width=\textwidth]{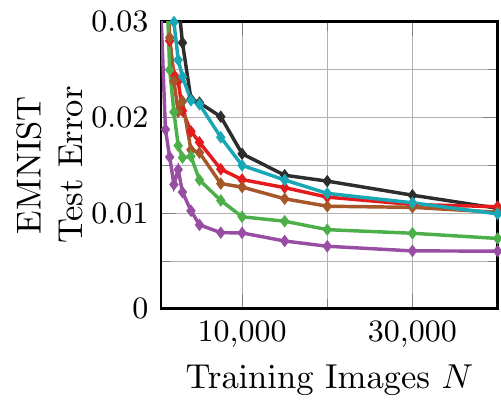}
    \end{subfigure}
    \begin{subfigure}{0.235\textwidth}
        \centering
        \includegraphics[width=\textwidth]{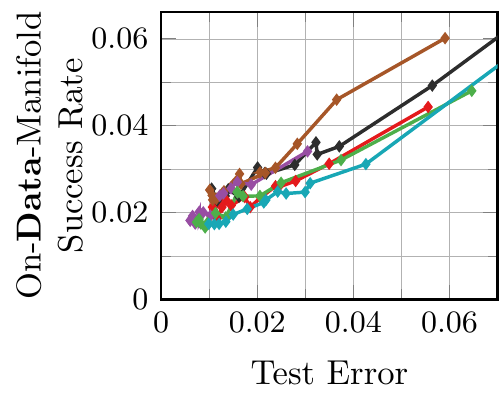}
    \end{subfigure}
    \\
    \fcolorbox{black!50}{white}{
        \begin{subfigure}{0.45\textwidth}
            \centering
            \includegraphics[width=1\textwidth]{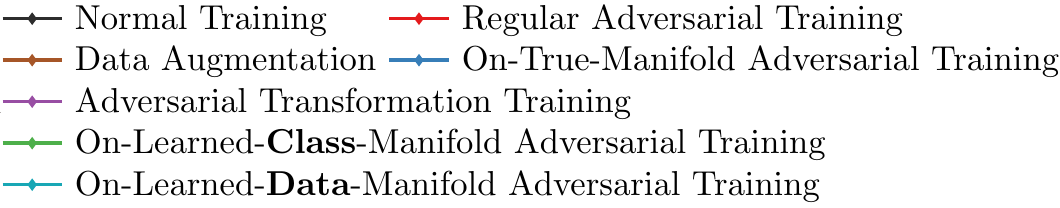}
        \end{subfigure}
    }
    \caption{Test error and on-data-manifold success rate on \Fonts and \MNIST. Using class-agnostic \VAEGANs, without clear class boundaries, on-manifold adversarial training looses its effectiveness -- the on-manifold adversarial examples cross the true class boundaries too often. The strong relationship between on-manifold robustness and generalization can still be confirmed.}
    \label{fig:appendix-data}
\end{figure}

In the context of Sections 3.3 and 3.4 of the main paper, we consider approximating the manifold using class-agnostic \VAEGANs. Instead of the class-conditionals $p(x|y)$ of the data distribution, the marginals $p(x)$ are approximated, \ie, images of different classes are embedded in the same latent space. Then, however, ensuring label invariance, as required by our definition of on-manifold adversarial examples, becomes difficult:
\begin{definition}[On-Manifold Adversarial Example]
    Given the data distribution $p$, an on-manifold adversarial example for $x$ with label $y$ is a perturbed version $\tilde{x}$ such that $f(\tilde{x}) \neq y$ but $p(y|\tilde{x}) > p(y'|\tilde{x}) \forall y' \neq y$.
    \label{def:appendix-on-manifold-adversarial-example}
\end{definition}
\noindent Therefore, we attempt to ensure \defref{def:appendix-on-manifold-adversarial-example} through a particularly small $L_{\infty}$-constraint on the perturbation, specifically $\|\zeta\|_{\infty} \leq \eta$ with $\eta = 0.1$ where $\zeta$ is the perturbation applied in the latent space. Still, as can be seen in \figref{fig:appendix-data-examples}, on-manifold adversarial examples might cross class boundaries, \ie, they change their actual label rendering them invalid according to our definition.

In \figref{fig:appendix-data}, we clearly distinguish between on-\emph{class}-manifold and on-\emph{data}-manifold adversarial training, corresponding to the used class-specific or -agnostic \VAEGANs. Robustness, however, is measured \wrt on-data-manifold adversarial examples.  As can be seen, the positive effect of on-manifold adversarial training diminishes when using on-data-manifold adversarial examples during training. Both, on \Fonts and \MNIST, generalization slightly decreases in comparison to normal training because adversarial examples are not useful for learning the task if label invariance cannot be ensured. When evaluating robustness against on-data-manifold adversarial examples, however, the relation of on-data-manifold robustness to generalization can clearly be seen. Overall, this shows that this relationship also extends to more general, less strict definitions of on-manifold adversarial examples.

\begin{figure}[t]
    \centering
    \vskip -0.4cm
    \begin{subfigure}[t]{0.235\textwidth}
        \centering
        \includegraphics[width=1\textwidth]{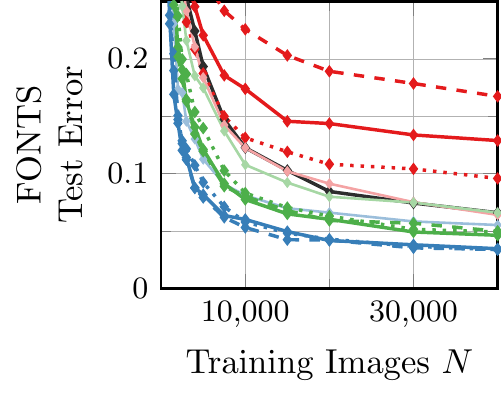}
    \end{subfigure}
    \begin{subfigure}[t]{0.235\textwidth}
        \centering
        \includegraphics[width=1\textwidth]{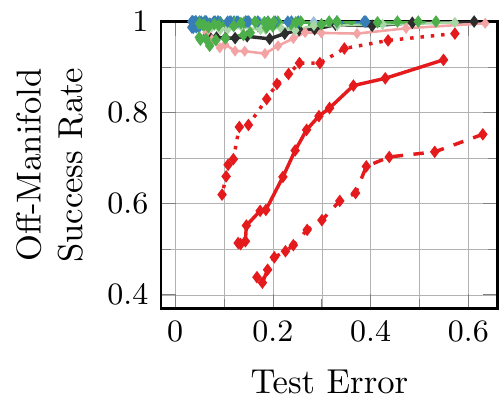}
    \end{subfigure}
    \\
    \begin{subfigure}[t]{0.235\textwidth}
        \centering
        \includegraphics[width=1\textwidth]{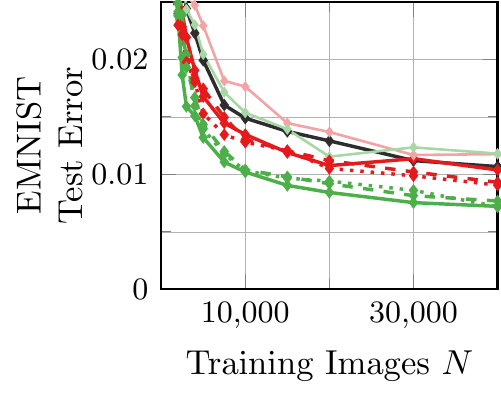}
    \end{subfigure}
    \begin{subfigure}[t]{0.235\textwidth}
        \centering
        \includegraphics[width=1\textwidth]{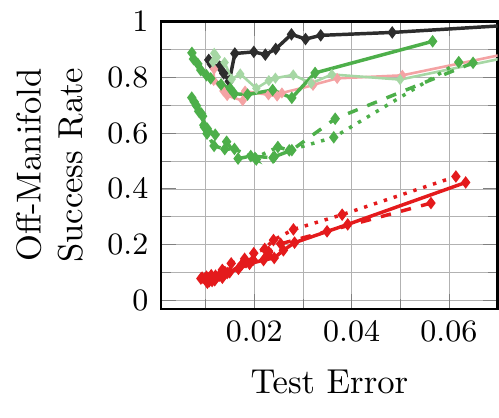}
    \end{subfigure}
    \\
    \fcolorbox{black!50}{white}{
        \begin{subfigure}{0.45\textwidth}
            \centering
            \includegraphics[width=0.7\textwidth]{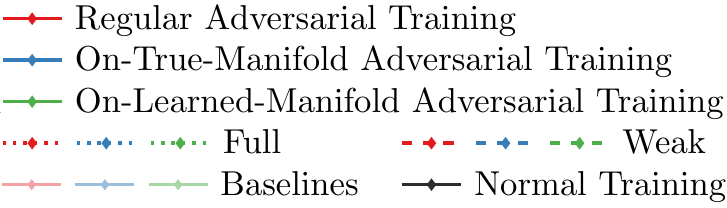}
        \end{subfigure}
    }
    \vskip -6px
    \caption{Adversarial training variants and baselines on \Fonts and \MNIST. For adversarial training, we consider the  \emph{full variant}, \ie, training on $100\%$ adversarial examples, and the \emph{weak variant}, \ie, stopping the inner optimization problem of \eqnref{eq:appendix-off-manifold-adversarial-training} as son as the first adversarial example is found. For regular adversarial training, the strength of the adversary determines the robustness-generalization trade-off; for on-manifold adversarial training, the ideal strength depends on the approximation quality of the used \VAEGANs.}
    \label{fig:appendix-baselines}
\end{figure}
\begin{figure}[t]
    \centering
    \vskip -0.2cm
    \includegraphics[width=0.4\textwidth]{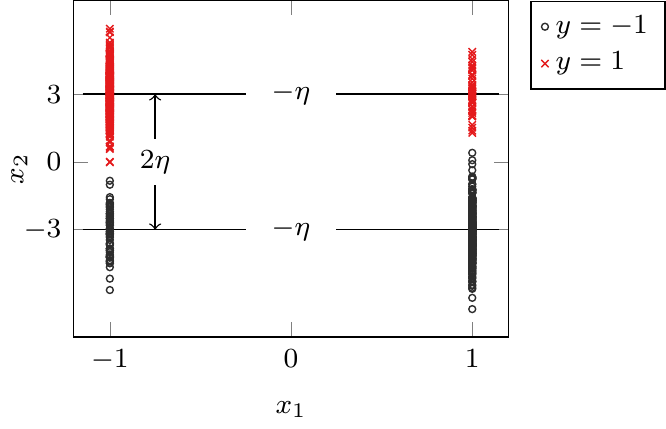}
    \vskip -6px
    \caption{Illustration of the toy dataset considered by Tsipras \etal in \cite{TsiprasARXIV2018} and defined in \eqnref{eq:appendix-tsipras}. For labels $y = 1$ and $y = -1$, the two-dimensional observations $x \in \{-1,1\}{\times}\mR$ are plotted. The first dimension, \ie, $x_1$, mirrors the label with probability $0.9$; the second dimension, \ie, $x_2$, is drawn from a Gaussian $\mathcal{N}(y3, I)$, \ie, $\eta$ from the text is $3$. As illustrated on the left, perturbing an observation $x$ with label $y = 1$ but $x_1 = -1$ by $2\eta = 6$ results in an adversarial example $\tilde{x}$ indistinguishable from observations with label $y = -1$.}
    \label{fig:appendix-tsipras}
\end{figure}

\section{Baselines and Adversarial Training Variants}
\label{sec:appendix-adversarian-training}

In the main paper, see Section 3.1, we consider the adversarial training variant by Madry \etal \cite{MadryICLR2018}, \ie,
\vskip -14px
\begin{align}
    \min_w \sum_{n = 1}^N \max_{\|\delta\|_{\infty} \leq \epsilon} \cL(f(x_n + \delta; w), y_n),
    \label{eq:appendix-off-manifold-adversarial-training}
\end{align}
\vskip -4px
\noindent where $f$ is the classifier with weights $w$, $\cL$ is the cross-entropy loss and $x_n$, $y_n$ are training images and labels. In contrast to \cite{MadryICLR2018}, we train on $50\%$ clean and $50\%$ adversarial examples \cite{SzegedyARXIV2013,GoodfellowARXIV2014}. The inner optimization problem is run for full $40$ iterations, as described in \secref{sec:appendix-attacks} without early stopping. Here, we additionally consider the \emph{full variant}, \ie, training on $100\%$ adversarial examples; and the \emph{weak variant}, \ie, stopping the inner optimization problem as soon as the label changes. Additionally, we consider random perturbations as baseline, \ie, choosing the perturbations $\delta$ uniformly at random without any optimization. The same variants and baselines apply to on-manifold adversarial training and adversarial transformation training.

In Section 3.6 of the main paper, we observed that different training strategies might exhibit different robustness-generalization characteristics. For example, regular adversarial training renders the learning problem harder: in addition to the actual task, the network has to learn (seemingly) random but adversarial noise directions leaving the manifold. In \figref{fig:appendix-baselines}, we first show that training on randomly perturbed examples (instead of adversarially perturbed ones) is not effective, neither in image space nor in latent space. This result highlights the difference between random and adversarial noise, as also discussed in \cite{FawziNIPS2016}. For regular adversarial training, the strength of the adversary primarily influences the robustness-generalization trade-off; for example, the weak variant increases generalization while reducing robustness. Note that this effect also depends on the difficulty of the task, \eg, \Fonts is considerably more difficult than \MNIST. For on-manifold adversarial training, in contrast, the different variants have very little effect; generalization is influenced only slightly, while regular robustness is -- as expected -- not influenced.

\section{Definition of Adversarial Examples}
\label{sec:appendix-adversarial-example}

Adversarial examples are assumed to be label-invariant, \ie, the actual, true label does not change. For images, this is usually enforced using a norm-constraint on the perturbation -- \eg, \cf \eqnref{eq:appendix-off-manifold-madry}; on other modalities, however, this norm-constraint might not be sufficient. In Section 3.3 of the main paper, we provide a definition for on-manifold adversarial examples based on the true, underlying data distribution -- as restated in \defref{def:appendix-on-manifold-adversarial-example}. Here, we use this definition to first discuss a simple and intuitive example before considering the theoretical argument of \cite{TsiprasARXIV2018}, claiming that robust \emph{and} accurate models are not possible on specific datasets; an argument in contradiction to our results

Let the observations $x$ and labels $y$ be drawn from a data distribution $p$, \ie, $x, y \sim p(x, y)$. Then, given a classifier $f$ we define adversarial examples as follows:
\begin{definition}[Adversarial Example]
    Given the data distribution $p$, an adversarial example for $x$ with label $y$ is a perturbed version $\tilde{x}$ such that $f(\tilde{x}) \neq y$ but $p(y|\tilde{x}) > p(y'|\tilde{x}) \forall y' \neq y$.
    \label{def:appendix-adversarial-example}
\end{definition}
\noindent In words, adversarial examples must not change the actual, true label \wrt the data distribution. Note that this definition is identical to \defref{def:appendix-on-manifold-adversarial-example} for on-manifold adversarial examples. For the following toy examples, however, the data distribution has non-zero probability on the whole domain or we only consider adversarial examples $\tilde{x}$ with $p(\tilde{x}) > 0$ such that \defref{def:appendix-adversarial-example} is well-defined. We leave a more general definition of adversarial examples for future work.

We illustrate \defref{def:appendix-adversarial-example} on an intuitive, binary classification task. Specifically, the classes $y = 1$ and $y = -1$ are uniformly distributed, \ie, $p(y = 1) = p(y = -1) = 0.5$ and observations are drawn from point masses on $0$ and $\epsilon$:
\vskip -14px
\begin{align}
p(x = 0|y=1) &= 1\\
p(x = \epsilon|y = -1) &= 1
\end{align}
\vskip -4px
\noindent  This problem is linearly separable for any $\epsilon > 0$; however, it seems that no classifier will be adversarially robust against perturbations of absolute value $\epsilon$. For simplicity, we consider the observation $x = 0$ with $y = 1$ and the adversarial example $\tilde{x} = x + \epsilon = \epsilon$. Then, verifying \defref{def:appendix-adversarial-example} yields a contradiction:
\vskip -14px
\begin{align}
    0 = p(y = 1 | x = \epsilon) \not> p(y = -1 | x = \epsilon) = 1.
\end{align}
\vskip -4px
\noindent  It turns out, $\tilde{x} = \epsilon$ is not a proper adversarial example. This example illustrates that an exact definition of adversarial examples, \eg, \defref{def:appendix-adversarial-example}, is essential to study the robustness of such toy datasets.

\subsection{Discussion of \cite{TsiprasARXIV2018}}

In \cite{TsiprasARXIV2018}, Tsipras \etal argue that there exists an inherent trade-off between regular robustness and generalization based on a slightly more complex toy example; we follow the notation in \cite{TsiprasARXIV2018}. Specifically, for labels $y = 1$ and $y = -1$ with $p(y = 1) = p(y = -1) = 0.5$, the observations $x \in \{-1,1\}{\times}\mR$ are drawn as follows\footnote{Note that, for simplicity and convenience, we consider the $2$-dimensional case; Tsipras \etal consider the general $D$-dimensional case, where $x_1$ remains unchanged and $x_2,\ldots, x_D$ are drawn from the corresponding Gaussian, \cf \eqref{eq:appendix-tsipras}.}:
\vskip -14px
\begin{align}
    \begin{split}
        p(x_1 | y) &= \begin{cases}p & \text{ if }x_1 = y\\1-p & \text{ if }x_1 = -y\end{cases},\\\quad p(x_2 | y) &= \mathcal{N}(x_2; y\eta, 1)
    \end{split}
    \label{eq:appendix-tsipras}
\end{align}
\vskip -4px
\noindent where $\eta$ defines the degree of overlapping between the two classes and $p \geq 0.5$. \figref{fig:appendix-tsipras} illustrates this dataset for $p = 0.9$ and $\eta = 3$. For a $L_{\infty}$-bounded adversary with $\epsilon \geq 2\eta$, Tsipras \etal show that no model can be both accurate and robust. Specifically, for $x$ with $y = 1$ but $x_1 = -1$ and $x_2 = \eta$, we consider replacing $x_2$ with $\tilde{x}_2 = x_2 - 2\eta = -\eta$, as considered in \cite{TsiprasARXIV2018}. However, this adversary does not produce proper adversarial examples according to our definition. Indeed,
\vskip -14px
\begin{align}
    \begin{split}
        p(y = 1 |& x = \tilde{x})\\
        &= p(y = 1 | x_1 = -1) \cdot p(y = 1 | x_2 = - \eta)\\
        &= (1 - p) \cdot \mathcal{N}(x_2 = - \eta; \eta, 1)\\
        &\not> p \cdot \mathcal{N}(x_2 = - \eta; -\eta, 1)\\
        &= p(y = -1 | x_1 = -1) \cdot p(y = -1 | x_2 = - \eta)\\
        &= p(y=-1 | x = \tilde{x})
    \end{split}
\end{align}
\vskip -4px
\noindent which contradicts our definition. Thus, in light of \defref{def:appendix-adversarial-example}, the suggested trade-off of Tsipras \etal is questionable. However, we note that this argument explicitly depends on our definition of proper and invalid adversarial examples, \ie, \defref{def:appendix-adversarial-example}; other definitions of adversarial examples or adversarial robustness, \eg, in the context of the adversarial loss defined in \cite{TsiprasARXIV2018}, may lead to different conclusions.
\end{appendix}

\end{document}